\documentclass[runningheads]{llncs}
\usepackage{graphicx}

\usepackage{tikz}
\usepackage{comment}
\usepackage{amsmath,amssymb} 
\usepackage{color}
\usepackage{graphicx}
\usepackage{makecell}
\usepackage{booktabs}
\usepackage{bm}
\usepackage{hyperref}
\usepackage[accsupp]{axessibility}  

\begin{document}
\pagestyle{headings}
\mainmatter
\def\ECCVSubNumber{4287}  

\title{Contributions of Shape, Texture, and Color in Visual Recognition} 

\titlerunning{Contributions of Shape, Texture, and Color in Visual Recognition}
%

\author{Yunhao Ge{$^*$} \and
Yao Xiao{$^*$}\and
Zhi Xu \and
Xingrui Wang \and
Laurent Itti
}
\authorrunning{Y. Ge and Y. Xiao et al.}
%
\institute{University of Southern California \\
\textcolor{magenta}{\url{https://github.com/gyhandy/Humanoid-Vision-Engine}}
}

\maketitle

\begin{abstract}

We investigate the contributions of three important features of the human visual system (HVS)~---~shape, texture, and color ~---~to object classification. We build a humanoid vision engine (HVE) that explicitly and separately computes shape, texture, and color features from images. The resulting feature vectors are then concatenated to support the final classification. We show that HVE can summarize and rank-order the contributions of the three features to object recognition. We use human experiments to confirm that both HVE and humans predominantly use some specific features to support the classification of specific classes (e.g., texture is the dominant feature to distinguish a zebra from other quadrupeds, both for humans and HVE). With the help of HVE, given any environment (dataset), we can summarize the most important features for the whole task (task-specific; e.g., color is the most important feature overall for classification with the CUB dataset), and for each class (class-specific; e.g., shape is the most important feature to recognize boats in the iLab-20M dataset). To demonstrate more usefulness of HVE, we use it to simulate the open-world zero-shot learning ability of humans with no attribute labeling. Finally, we show that HVE can also simulate human imagination ability with the combination of different features. We will open-source the HVE engine and corresponding datasets.
\end{abstract}

\def\thefootnote{*}\footnotetext{Yunhao Ge and Yao Xiao contributed equally}\def\thefootnote{\arabic{footnote}}

\section{Introduction}
\begin{figure}[t]
\begin{center}
\includegraphics[width=\linewidth]{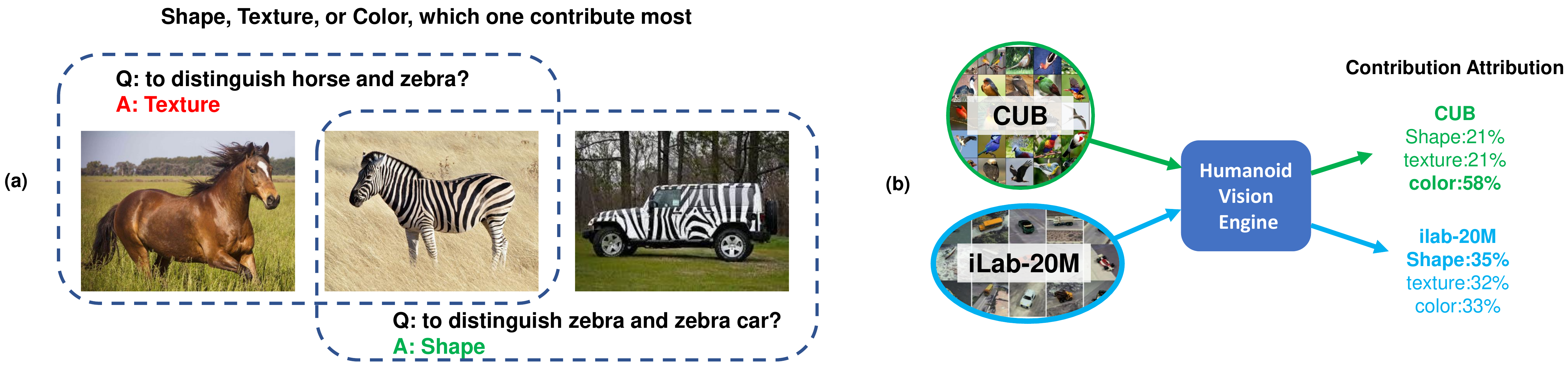}

\end{center}
   \caption{Left: Contributions of Shape, Texture, and Color may be different among different scenarios/tasks. Here, texture is most important to distinguish zebra from horse, but shape is most important for zebra vs.~zebra car. Right: Humanoid Vision Engine takes dataset as input and summarizes how shape, texture, and color contribute to the given recognition task in a pure learning manner (E.g., In ImageNet classification, shape is the most discriminative feature and contributes most to visual recognition). }
\label{fig:motivation_figure}
\end{figure}

The human vision system (HVS) is the gold standard for many current computer vision algorithms, on various challenging tasks: zero/few-shot learning ~\cite{PalatucciPHM09,lampert2009learning,sung2018learning,rahman2018unified,SnellSZ17}, meta-learning~\cite{andrychowicz2016learning,KhodadadehBS19}, continual learning~\cite{schlimmer1986case,thrun1995lifelong,wen2021beneficial}, novel view imagination~\cite{NEURIPS2018_92cc2275,GeAXI21}, etc.
Understanding the mechanism, function, and decision pipeline of HVS becomes more and more important. 
The vision systems of humans and other primates are highly differentiated. Although HVS provides us a unified image of the world around us, this picture has multiple facets or features, like shape, depth, motion, color, texture, etc.~\cite{gazzaniga2006cognitive,grill2004human}. To understand the contributions of the most important three features~---~shape, texture, and color~---~in visual recognition, some research compares the HVS with an artificial convolutional Neural Network (CNN). A widely accepted intuition about the success of CNNs on perceptual tasks is that CNNs are the most predictive models for the human ventral stream object recognition ~\cite{perceptiondeep,yamins2014performance}. 
To understand which feature is more important for CNN-based recognition, recent paper shows promising results: ImageNet-trained CNNs are biased towards texture while increasing shape bias improves accuracy and robustness \cite{li2020shapetexture}. 

Due to the superb success of HVS on various complex tasks ~\cite{PalatucciPHM09,andrychowicz2016learning,schlimmer1986case,NEURIPS2018_92cc2275,ge2021peek}, human bias may also represent the most efficient way to solve vision tasks. And it is likely task-dependent (Fig.~\ref{fig:motivation_figure}). 
Here, inspired by HVS, we wish to find a general way to understand how shape, texture, and color contribute to a recognition task by pure data-driven learning.
The summarized feature contribution is important both for the deep learning community (guide the design of accuracy-driven models \cite{li2020shapetexture,GeirhosRMBWB19,gatys2017texture,BrendelB19}) and for the neuroscience community (understanding the contributions or biases in human visual recognition) \cite{OLIVA2000176,doi:10.1073/pnas.1015666108}.

It has been shown by neuroscientists that there are separate neural pathways to process these different visual features in primates~\cite{amir1993cortical,deyoe1996mapping}. Among the many kinds of features crucial to  visual recognition in humans, the shape property is the one that we primarily rely on in static object recognition ~\cite{gazzaniga2006cognitive}. Meanwhile, some previous studies show that surface-based cues also play a key role in our vision system. For example, \cite{gegenfurtner2000sensory} shows that scene recognition is faster for color images compared with grayscale ones and \cite{puce1996differential,peuskens2004attention}  found a special region in our brain to analyze textures. In summary, \cite{cant2008independent,cant2007attention} propose that shape, color and texture are three separate components to identify an object. 



To better understand the task-dependent contributions of these features, we build a Humanoid Vision Engine (HVE) to simulate HVS by explicitly and separately computing shape, texture, and color features to support image classification in an objective learning pipeline.
HVE has the following key contributions: 
(1) Inspired by the specialist separation of the human brain on different features \cite{amir1993cortical,deyoe1996mapping}, for each feature among shape, texture, and color, we design a specific feature extraction pipeline and representation learning model. 
(2) To summarize the contribution of features by end-to-end learning, we design an interpretable humanoid Neural Network (HNN) that aggregates the learned representation of three features and achieves object recognition, while  also showing the contribution of each feature during decision.  
(3) We use HVE to analyze the contribution of shape, texture, and color on three different tasks subsampled from ImageNet. 
We conduct human experiments on the same tasks and show that both HVE and humans predominantly use some specific features to support object recognition of specific classes. 
(4) We use HVE to explore the contribution, relationship, and interaction of shape, texture, and color in visual recognition. Given any environment (dataset), HVE can summarize the most important features (among shape, texture, and color) for the whole task (task-specific) and for each class (class-specific). To the best of our knowledge, we provide the first fully objective, data-driven, and indeed first-order, quantitative measure of the respective contributions.
(5) HVE can help guide accuracy-driven model design and performs as an evaluation metric for model bias. For more applications, we use HVE to simulate the open-world zero-shot learning ability of humans which needs no attribute labels. HVE can also simulate human imagination ability across features. We open-source the HVE engine and corresponding dataset.


\section{Related Works}

In recent years, more and more researchers focus on the interpretability and generalization of computer vision models like CNN~\cite{SimonyanZ14a,he2016deep} and vision transformer~\cite{DosovitskiyB0WZ21}. For CNN, many researchers try to explore what kind of information is most important for models to recognize objects. Some paper show that CNNs trained on the ImageNet are more sensitive to texture information~\cite{GeirhosRMBWB19,gatys2017texture,BrendelB19}. But these works fail to quantitatively explain the contribution of shape, texture, color as different features, comprehensively in various datasets and situations. While most recent studies focus on the bias of Neural Networks, exploring the bias of humans or a humanoid learning manner is still under-explored and inspiring.

Besides, many researchers contribute to the generalization of computer vision models and focus on zero/few-shot learning~\cite{PalatucciPHM09,lampert2009learning,vinyals2016matching,SnellSZ17,ge2021towards,Cheng2021}, novel view imagination~\cite{NEURIPS2018_92cc2275,GeAXI21,ge2022dall}, open-world recognition~\cite{bendale2015towards,joseph2021towards,jain2014multi}, etc. Some of them tackled these problems by feature learning --- representing an object by different features, and made significant progress in this area~\cite{tokmakov2019learning,PrabhudesaiLPTH21,NEURIPS2018_92cc2275}. But, there still lacks a clear definition of what these properties look like or a uniform design of a system that can do humanoid tasks like generalized recognition and imagination.


\section{Humanoid Vision Engine}
\begin{figure*}[h]
\begin{center}
\includegraphics[width=\linewidth]{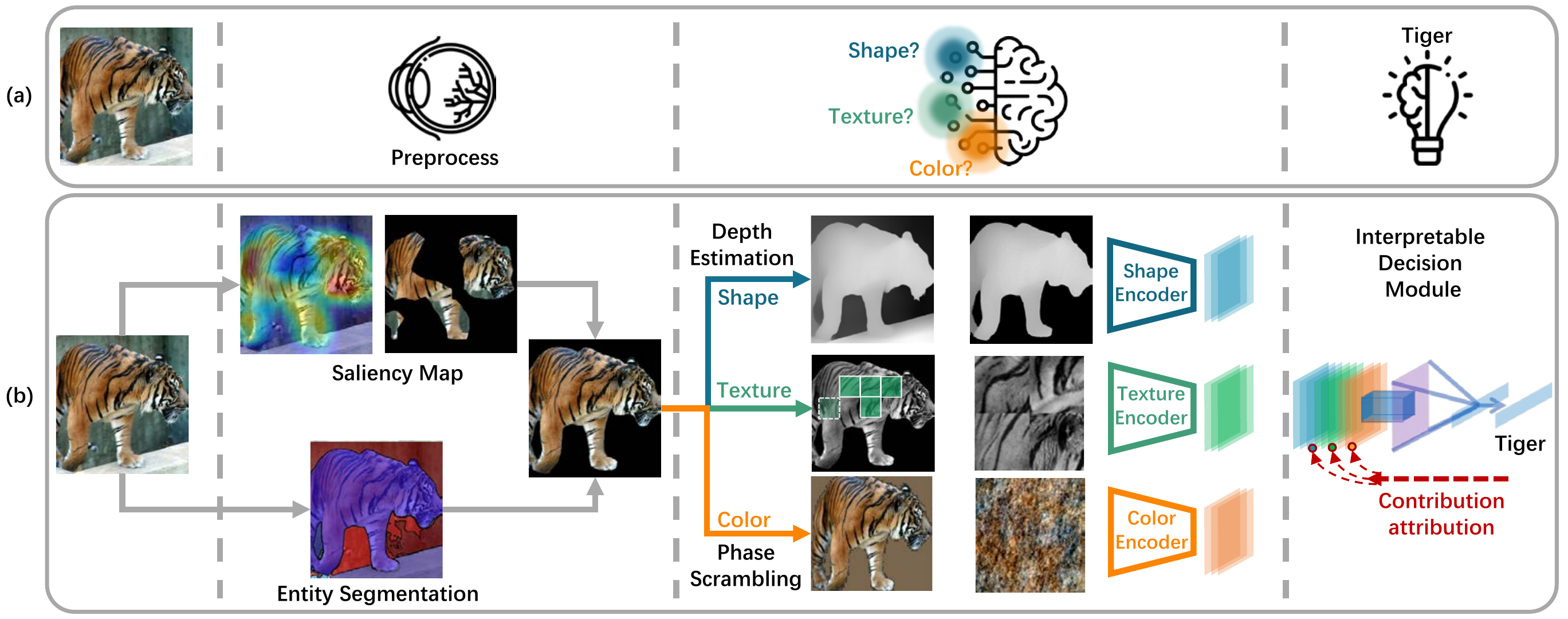}
\end{center}
   \caption{Pipeline for humanoid vision engine (HVE). (a) shows how will humans' vision system deal with an image. After humans' eyes perceive the object, the different parts of the brain will be activated. The human brain will organize and summarize that information to get a conclusion. (b) shows how we design HVE to correspond to each part of the human's vision system.}
\label{fig:pipeline}
\end{figure*}

The goal of the humanoid vision engine (HVE) is to summarize the contribution of shape, texture, and color in a given task (dataset) by separately computing the three features to support image classification, similar to the way that humans recognize objects. 
During the pipeline and model design, we borrow the findings of neuroscience researchers on the structure, mechanism, and function of HVS \cite{amir1993cortical,deyoe1996mapping,gazzaniga2006cognitive,gegenfurtner2000sensory,puce1996differential,peuskens2004attention}. We use end-to-end learning with backpropagation to simulate the learning process of humans and to summarize the contribution of shape, texture, and color. The advantage of end-to-end training is that we can avoid introducing human bias, which may influence the objective of contribution attribution (e.g., we do not introduce handcrafted elementary shapes as done in Recognition by Components \cite{biederman1987recognition}). We only use data-driven learning, a straightforward way to understand the contribution of each feature from effectiveness perspective, and we can easily generalize HVE to different tasks (datasets). As shown in Fig.~\ref{fig:pipeline}, HVE consists of (1) \textbf{a humanoid image preprocessing pipeline}, (2) \textbf{feature representation} for shape, texture, and color, and (3) \textbf{a humanoid neural network} that aggregates the representation of each feature and achieves interpretable object recognition.

\subsection{Humanoid Image Preprocessing and Feature Extraction}

As shown in Fig.\ref{fig:pipeline} (a), humans (or primates) can localize an object intuitively in a complex scene before we recognize what it is \cite{julesz1964binocular}. Also, there are different types of cells or receptors in our primary visual cortex extracting specific information (like color, shape, texture, shading, motion, etc) information from the image \cite{gazzaniga2006cognitive}. In our HVE (Fig.~\ref{fig:pipeline} (b)), for an input raw image $I\in \mathbb{R}^{H\times W \times C}$, we first parse the object from the scene as preprocessing and then extract our defined shape, texture, and color features $I_s, I_t, I_c$, 
for the following humanoid neural network. 

\subsubsection{Image Parsing and Foreground Identification.}




As shown in the preprocessing part of Fig.\ref{fig:pipeline} (b), we use the entity segmentation method \cite{qi2021open} to simulate the process of parsing objects  from a scene in our brain. Entity segmentation is an open-world model and can segment the object from the image without labels. This method aligns with human behavior, which can (at least in some cases; e.g., autostereograms~\cite{julesz1964binocular}) segment an object without deciding what it is. After we get the segmentation of the image, we use a pre-trained CNN and GradCam~\cite{selvaraju2017grad} to find the foreground object among all masks. (More details in Supp.)

We design three different feature extractors after identifying the foreground object segment: shape extractor, texture extractor, and color extractor, similar to the separate neural pathways in the human brain which focus on specific property~\cite{amir1993cortical,deyoe1996mapping}. The three extractors focus only on the corresponding features, and the extracted features, shape $I_s$, texture $I_t$, and color $I_c$, are disentangled from each other.

\subsubsection{Shape Feature Extractor}


For the shape extractor, we want to keep both 2D and 3D shape information while eliminating the information of texture and color. We first use a 3D depth prediction model \cite{Ranftl2020,Ranftl2021} to obtain the 3D depth information of the whole image. After element-wise multiplying the 3D depth estimation and 2D mask of the object, we obtain our shape feature $I_s\in \mathbb{R}^{H\times W}$. We can notice that this feature only contains 2D shape and 3D structural information (the 3D depth) and without color or texture information (Fig.~\ref{fig:pipeline}(b)). 


\subsubsection{Texture Feature Extractor}

\begin{figure}[h]
\begin{center}
\includegraphics[width=0.8\linewidth]{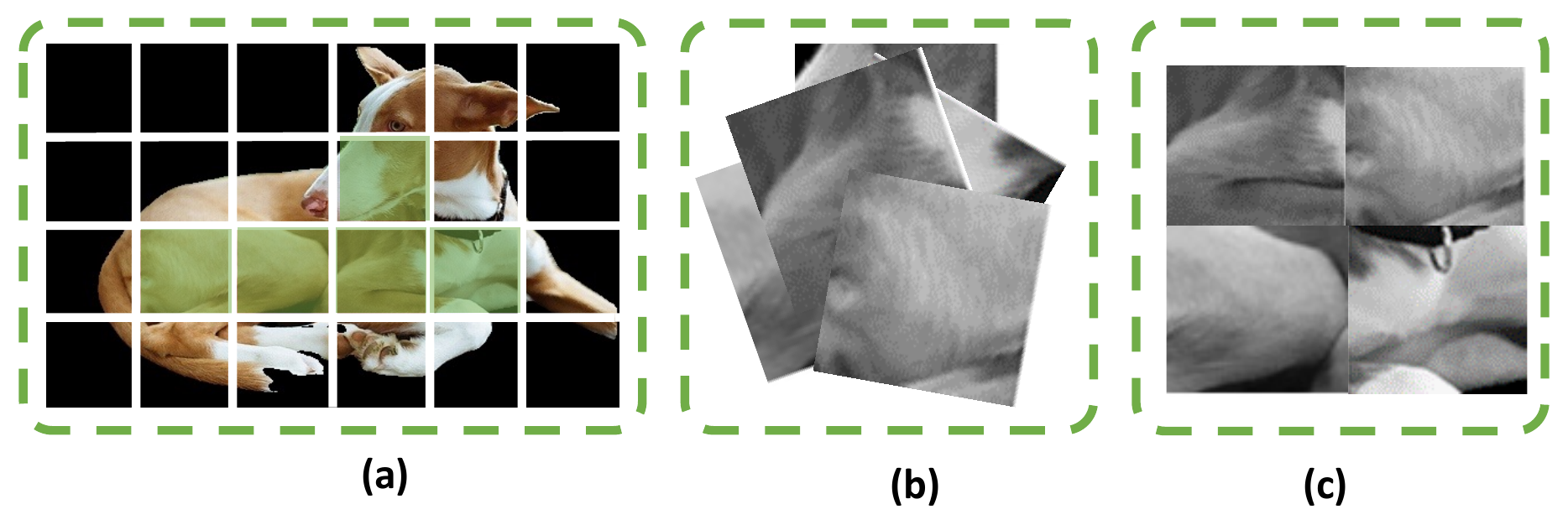}
\end{center}
   \caption{Pipeline for extracting texture feature: (a) Crop images and compute the overlap ratio between 2D mask and patches. Patches with overlap $>0.99$ are shown in a green shade. (b) add the valid patches to the patch pool. (c) randomly choose 4 patches from the pool and concatenate them to obtain a texture image $I_t$.}
  \vspace{-15pt}
\label{fig:texture}
\end{figure}

In texture extractor, we want to keep both local and global texture information while eliminating shape and color information. Fig.~\ref{fig:texture} visualizes the extraction process. First, to remove the color information, we convert the RGB object segmentation to a grayscale image. Next, we cut this image into several square patches with an adaptive strategy (the patch size and location are adaptive with object sizes to cover more texture information). If the overlap ratio between the patch and the original 2D object segment is larger than a threshold $\tau$, we  add that patch to a patch pool (we set $\tau$ to be 0.99 in our experiments, which means the over 99\% of the area of the patch belongs to the object). Since we want to extract both local (one patch) and global (whole image) texture information, we randomly select 4 patches from the patch pool and concatenate them into a new texture image ($I_t$). (More details in Supp.)

\subsubsection{Color Feature Extractor}
We use two methods to represent the color feature for a given image $I$. The first method is phase scrambling, which is popular in psychophysics and in signal processing~\cite{oppenheim1981importance,thomson1999visual}. Phase scrambling transforms the image into the frequency domain using the fast Fourier transform (FFT). In the frequency domain, the phase of the signal is then randomly scrambled, which destroys shape information while preserving color statistics. Then we use IFFT to transfer back to image space and get $I_c \in \mathbb{R}^{H\times W \times C}$. $I_c$ and $I$ have the same distribution of pixel color values (Fig.~\ref{fig:pipeline}(b)). 

We also used simple color histograms (see suppl.) as an alternative, but the results were not as good, hence we focus here on the phase scrambling approach for color representation. 

\subsection{Humanoid Neural Network}
\label{section:HNN}
After preprocessing, we have three features, i.e. shape $I_s$, texture $I_t$, color $I_c$ of an input image $I$.
To simulate the separate neural pathways in humans' brains for different feature information \cite{amir1993cortical,deyoe1996mapping}, we design three feature representation encoders for shape, texture, and color, respectively. Shape feature encoder $\mathnormal{E}_{s}$ takes a 3D shape feature $I_s$ as input and outputs the shape representation ($V_s = \mathnormal{E}_{s}(I_s) $). 
Similarly, texture encoder $\mathnormal{E}_{t}$ and color encoder $\mathnormal{E}_{c}$ take the texture patch image $I_t$ or color phase scrambled image $I_c$ as input, after embedded by $\mathnormal{E}_{t}$ (or $\mathnormal{E}_{c}$), we get the texture feature $V_t$ and color feature $V_c$.




We use ResNet-18~\cite{he2016deep} as the backbone for all feature encoders to project the three types of features to the corresponding well-separated embedding spaces. It is hard to define the ground-truth label of the  distance between features. Given that the objects from the same class are relatively consistent in shape, texture, and color, the encoders can be trained in the classification problem independently instead, with the supervision of class labels. After training our encoders as classifiers, the feature map of the last convolutional layer will serve as the final feature representation. (More details in the appendix.)



To aggregate separated feature representations and conduct object recognition, we freeze the three encoders and train a contribution interpretable aggregation module $\text{Aggr}_{\theta}$, which is composed of two fully-connected layers (Fig.~\ref{fig:pipeline} (b) right). (More exploration in Appendix B.3 ). We concatenate $V_s, V_t, V_c$ and send it to $\text{Aggr}_{\theta}$. The output is denoted as $p\in\mathbb{R}^{n}$, where $n$ is the number of classes. 
\begin{align}
p=\text{Aggr}_{\theta}\left(\text{concat}(V_s, V_t, V_c)\right).
\end{align}


We also propose a gradient-based {\em contribution attribution} method to interpret the contributions of shape, texture, and color to the classification decision, respectively.
Take the shape feature as an example, given a prediction $p$ and the probability of class $k$, namely $p^k$, we compute the gradient of $p^{k}$ with respect to the shape feature $V^{s}$. We define the gradient as shape importance weights $\alpha^{k}_{s}$:
\begin{align}
\alpha^{k}_{s}=\frac{\partial p^{k}}{\partial V_{s}} \quad 
\alpha^{k}_{t}=\frac{\partial p^{k}}{\partial V_{t}} \quad
\alpha^{k}_{c}=\frac{\partial p^{k}}{\partial V_{c}}. \quad
\end{align}

Then we can calculate element-wise product between $V_s$ and $\alpha^{k}_{s}$ to get the final shape contribution $S_s^k$. In other words, $S_s^k$ represents the ``contribution'' of shape feature to classifying this image as class $k$. 
\begin{align}
S^{k}_{s}=\text{ReLU}\left(\sum \alpha_{s}^{k} V_{s}\right).
\end{align}

We can do the same thing to get texture contribution $S^{k}_{t}$ and color contribution $S^{k}_{c}$. After getting the feature contributions for each image, we can calculate the average value of all images in this class to assign feature contributions to each class (class-specific bias) and the average value of all classes to assign feature contributions to the whole dataset (task-specific bias).

\section{Experiments}

In this section, 
we first show the effectiveness of feature encoders on representation learning (Sec.~\ref{sec:4.1}); then we show the contribution interpretation performance of Humanoid NN on different feature-biased datasets in ImageNet (Sec.~\ref{sec:4.2}); 
We use human experiments to confirm that both HVE and humans predominantly use some specific features to support the classification of specific classes (Sec.~\ref{sec:4.3});
Then we use HVE to summarize the contribution of shape, texture, and color on different datasets (CUB\cite{wah2011caltech} and iLab-20M\cite{borji2016ilab}) (Sec.~\ref{sec:4.4}).

\subsection{Effectiveness of Feature Encoders}
\label{sec:4.1}

\begin{figure}[t]
\begin{center}
\includegraphics[width=\linewidth]{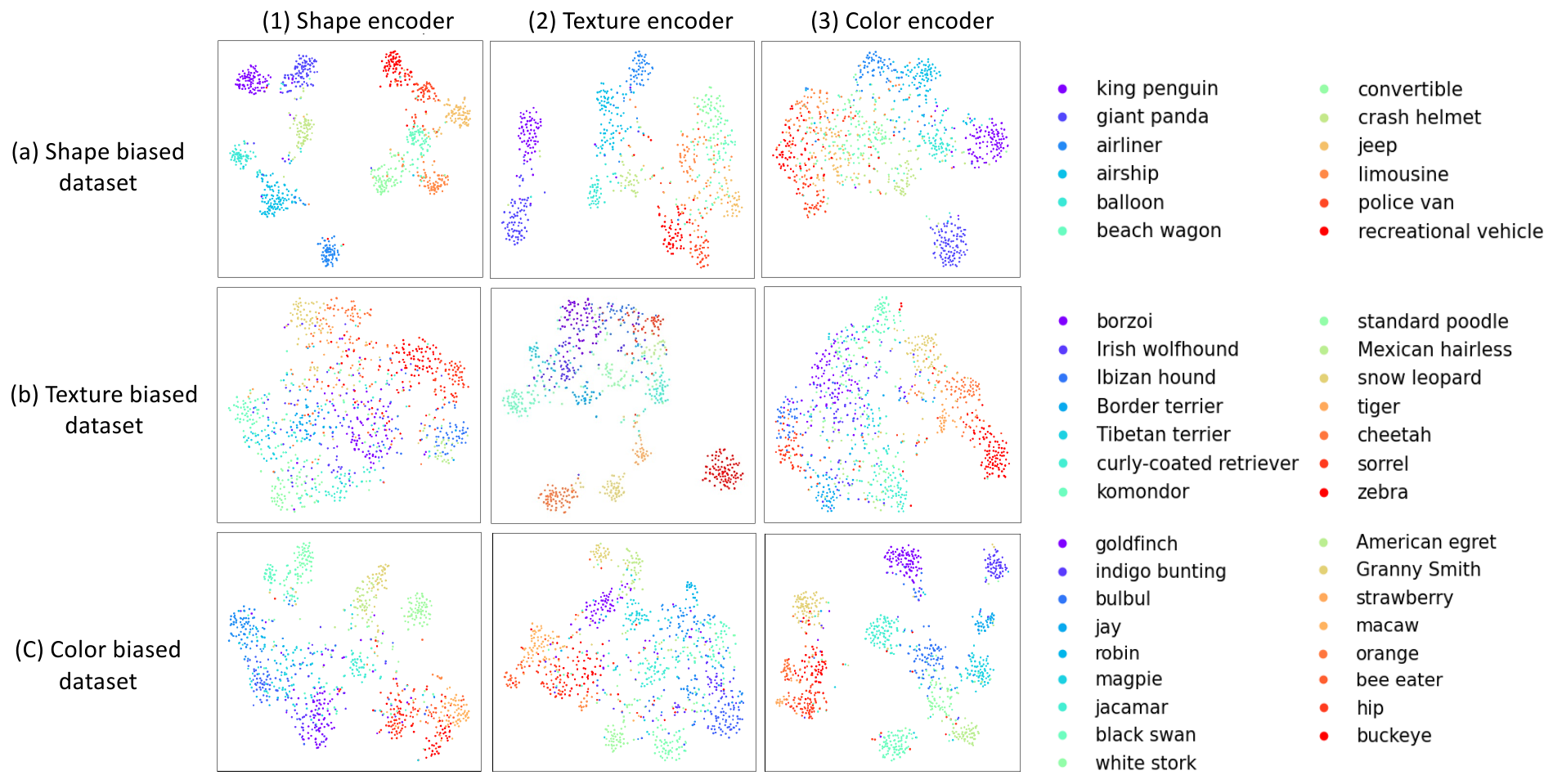}
\end{center}
   \caption{T-SNE results of feature encoders on their corresponding biased datasets}
\label{fig:dataset_tsne}
\end{figure}

To show that our three feature encoders focus on embedding their corresponding sensitive features,
we handcrafted three subsets of ImageNet \cite{krizhevsky2012imagenet}: shape-biased dataset ($D_{\text{shape}}$), texture-biased dataset ($D_{\text{texture}}$), and color-biased dataset ($D_{\text{color}}$). {\bf Shape-biased dataset} containing 12 classes, where the classes were chosen which intuitively are strongly determined by shape (e.g., vehicles are defined by shape more than color).  {\bf Texture-biased dataset} uses 14 classes which we believed are more strongly determined by texture. {\bf Color-biased dataset} includes 17 classes. The intuition of class selection of all three datasets will be verified by our results in Table~\ref{table:dataset-acc} with further illustration in Sec.~\ref{sec:4.2}. All these datasets use 80\% / 20\% of training and testing data (each class with around 800 training images and 200 testing images; images are randomly selected in each class). The details of classes contained in our biased datasets are shown in Fig.~\ref{fig:dataset_tsne}.

If our feature extractors actually learned their corresponding features with {\em feature-constructive} latent spaces, their T-SNE results will show clear clusters in the feature-biased datasets. ``Bias" here means we can classify the objects based on the biased feature easily, but it is more difficult to make decisions based on the other two features. As shown in Fig.~\ref{fig:motivation_figure}, distinguishing horse and zebra is a texture-biased task, while zebra vs.\ zebra car is a shape-biased task.

After pre-processing the original images and getting their feature images, we input the feature images into feature encoders and get the T-SNE results shown in Fig.~\ref{fig:dataset_tsne}. Each row represents one feature-biased dataset and each column is bounded with one feature encoder, each image shows the results of one combination. For instance, row (a) shows that the 12 classes in our shape-biased dataset are well separated in the embedding space of our shape encoder (column 1), but they are not separated in the embedding space of the texture (column 2) or color (column 3) encoders. Likewise color and texture encoders, their T-SNE results are separated perfectly on corresponding datasets (diagonal) but not as well on others' datasets (off-diagonal), which shows that our feature encoders are predominantly sensitive to the corresponding features.
\begin{table}[t]

\scriptsize
\centering
\caption{The ``original" column means the accuracy of Resnet18 on the original images is our upper bound. ``shape", ``texture" and ``color" columns represent the accuracy of feature nets. The ``all" column shows results of the Humanoid Neural Network that combines the 3 feature nets. It approaches the upper bound, suggesting that the split into 3 feature nets preserved most information needed for image classification.}

\begin{tabular}{c |c  c  c  c  c}
    \toprule
    accuracy & \bf{original} & shape & texture & color & \bf{all}\\
    \midrule
     Shape biased dataset & \bf{97\%} & \bf{90\%} & 84\% & 71\% & \bf{95\%} \\
     Texture biased dataset & \bf{96\%} & 64\% & \bf{81\%} & 65\% & \bf{91\%}  \\
     Color biased dataset & \bf{95\%} & 70\% & 73\% & \bf{82\%} &\bf{92\%}\\
     \bottomrule
\end{tabular}
\label{table:dataset-acc}
\end{table}

\subsection{Effectiveness of Humanoid Neural Network}
\label{sec:4.2}

We can use feature encoders to serve as classifiers after adding fully-connected layers. As these classifiers classify images based on corresponding feature representation, we call them \textit{feature nets}. We tested the accuracy of feature nets on these three biased datasets. As shown in Table~\ref{table:dataset-acc}, a ResNet-18 trained on the original segmented images (without explicit separated features, e.g. Fig.~\ref{fig:pipeline} (b) tiger without background)
provided an upper bound for the task. We find that feature net consistently obtains the best performance on their own biased dataset (e.g., on the shape-biased dataset, shape net classification performance is better than that of the color net or texture net). If we combine these three feature nets with the interpretable aggregation module, the classification accuracy is very close to the upper bound, which means our vision system can classify images based on these three features almost as well as based on the full original color images. This demonstrates that we can obtain most information of original images by our feature nets, and our aggregation and interpretable decision module actually learned how to combine those three features by end-to-end learning.

Table~\ref{table:biased-contributes} shows the quantitative contribution summary results of Humanoid NN (Sec.~\ref{section:HNN}). For task-specific bias on each dataset, shape plays a dominant role in shape-biased tasks, and texture, color also contribute most to their related biased tasks. 

\begin{table}
  \scriptsize
  \caption{Contributions of features for different biased datasets.}
  \centering
\begin{tabular}{c |c  c  c}
    \toprule
     contribution & shape ratio & texture ratio & color ratio\\
    \midrule
     Shape biased dataset & \bf{47\%} & 34\% & 19\%\\
     Texture biased dataset & 5\% & \bf{65\%} & 30\%\\
     Color biased dataset & 11\% & 19\% & \bf{70\%}\\
     \bottomrule
\end{tabular}
\label{table:biased-contributes}
\end{table}



\subsection{Human Experiments}
\label{sec:4.3}

Intuitively, we expect that humans may rely on different features to classify different objects (Fig.~\ref{fig:motivation_figure}). To show this, we designed human experiments that asked participants to classify reduced images with only shape, texture, or color features. If an object is mainly recognizable based on shape for humans, we could then check whether it is also the same for HVE, and likewise for color and texture.

\begin{figure}[h]
\begin{center}
\includegraphics[width=\linewidth]{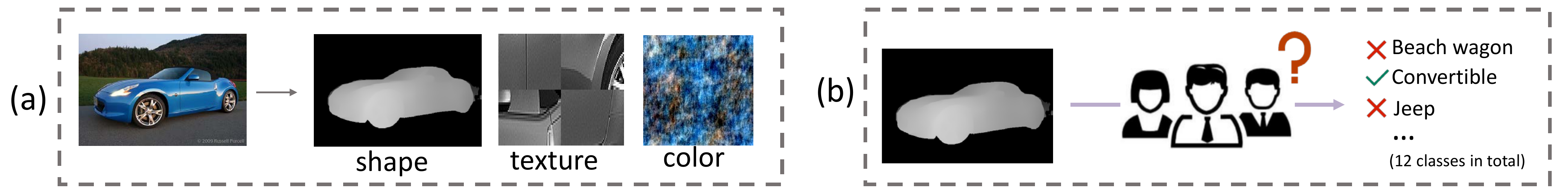}
\end{center}
    \caption{Sample question for the human experiment. (a) A test image (left) is first converted into shape, color, and texture images using our feature extractors. (b) On a given trial, human participants are presented with one shape, color, or texture image, along with 2 reference images for each class in the corresponding dataset (not shown here, see suppl. for a screenshot of an experiment trial). Participants are asked to guess the correct object class from the feature image.}
\label{fig:sample_question}
\end{figure}

\subsubsection{Experiments Design}

As shown in Table.~\ref{table:dataset-acc}, the three datasets (Fig.~\ref{fig:dataset_tsne}) have a clear bias towards corresponding features. We asked the participants to classify objects in the corresponding dataset based on one single feature image computed by one of our feature extractors (Fig.~\ref{fig:sample_question}). The reduced image contained only shape, texture, or color information. To make sure that the participants understood the class definitions well, we also showed them two example images of each class at the bottom of the screen with the assigned class label written below (for instance, some participants may not have been familiar with what the "beach wagon" ImageNet class is). Participants were asked to choose the correct class label for the reduced image (from 12/14/17 classes in shape/texture/color datasets).

\subsubsection{Human Performance Results}


The results here are based on 3270 trials, 109 participants. We publicly posted our questionnaires on the internet and sent the questionnaire link to the students and machine learning researchers in different universities. The accuracy for different feature questions on different biased datasets can be seen in Table~\ref{table-human-acc}. Human performance is quite similar to our feature nets' performance (compare Table~\ref{table:dataset-acc} with Table~\ref{table-human-acc}). On the shape-biased dataset, both human and feature nets attain the highest accuracy with shape. The same for the color and texture biased datasets. Thus, both HVE and humans predominantly use some specific features to support object recognition of specific classes. Interestingly, humans can perform not badly on all three biased datasets with shape feature.

\begin{table}
  \small
\caption{Humans' recognition accuracy of different feature images on different biased datasets}

  \centering
\begin{tabular}{c |c  c  c}
    \toprule
    accuracy & shape & texture & color\\
    \midrule
    
     Shape biased dataset & \bf{90.0\%} & 49.0\% & 16.8\% \\
     Texture biased dataset & 33.1\% & \bf{40.0\%} & 11.1\%  \\
     Color biased dataset & 32.3\% & 19.7\% &\bf{46.5\%}\\
     \bottomrule
\end{tabular}

\label{table-human-acc}
\end{table}



 
  

\begin{table}

\setlength\tabcolsep{3pt}
  \scriptsize
\caption{class-specific bias for each class in iLab-20M}

  \centering
\begin{tabular}{c | c  c  c  c  c  c c c  c c }
    \toprule
    ratio & boat& bus & car & mil & monster & pickup & semi & tank & train & van\\
    \midrule \midrule
shape & \bf{40\%} & 35\% & \bf{44\%} & 18\% & 36\% & 28\%  & \bf{40\%} & 36\% & 31\% & \bf{40\%}\\
 
texture & 32\% & 31\% & \bf{40\%} & 30\% & 34\% & 20\%  & 31\% & 32\% & 34\% & 27\%\\
  
color & 28\% & 34\% & 16\% & \bf{52\%} & 30\% & \bf{53\%}  & 29\% & 32\% & 35\% & 33\%\\
  \bottomrule
\end{tabular}
\label{table:ilab-local-bias}
\end{table}

\subsection{Contributions Attribution in Different Tasks}
\label{sec:4.4}
\begin{figure}
\begin{center}
\includegraphics[width=\linewidth]{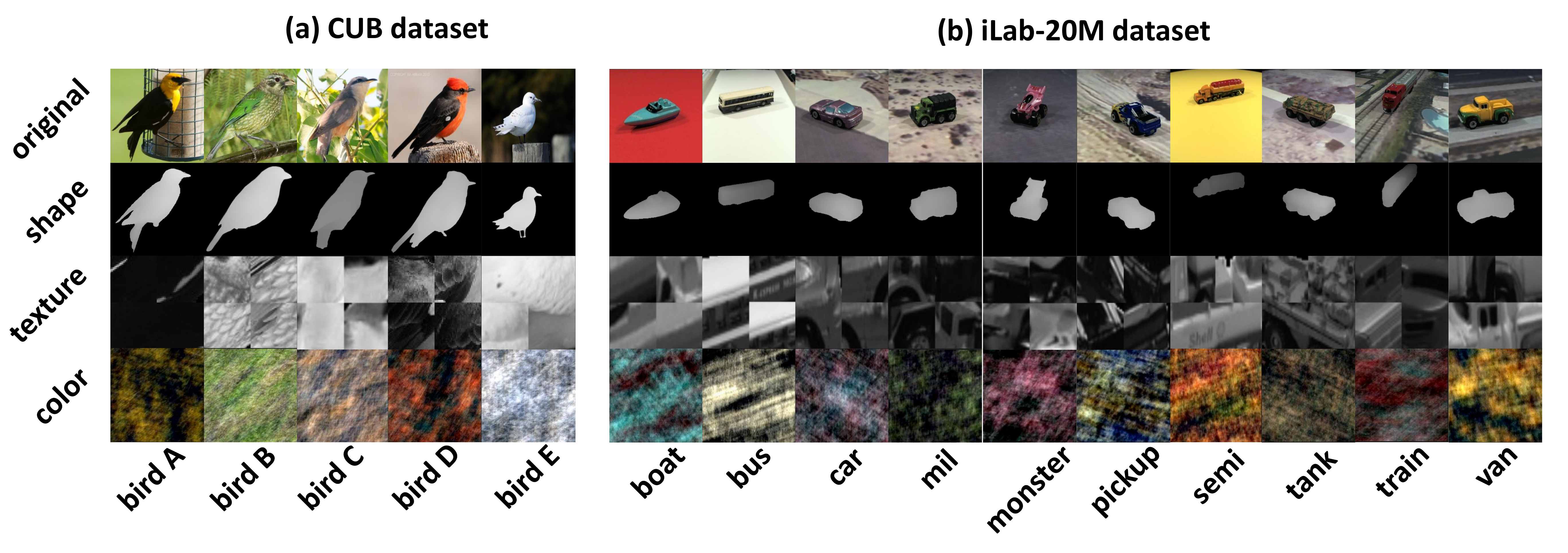}
\end{center}
\caption{Overview of CUB and iLab-20M dataset}
   
\label{fig:overview_dataset}
\end{figure}

With our vision system, we can summarize the task-specific bias and class-specific bias for any dataset. This enables several applications: (1) Guide accuracy-driven model design; e.g.,\cite{li2020shapetexture,GeirhosRMBWB19,gatys2017texture,BrendelB19} use dataset bias to guide the model design. Our method provides objective summarization of dataset bias. (2) Evaluation metric for model bias. Our method can help
correct an initially wrong model bias on some datasets (e.g., that most CNN trained on ImageNet are texture biased \cite{GeirhosRMBWB19,li2020shapetexture}). (3) Substitute human intuition to obtain more objective summarization with end-to-end learning.
We implemented the biased summarization experiments on two datasets, CUB \cite{wah2011caltech} and iLab-20M \cite{borji2016ilab}. Fig.~\ref{fig:motivation_figure}(b) shows the task-specific biased results. The results of CUB are reasonable since CUB is a dataset of birds, which means all the classes in CUB have  a similar shape with feather textures, hence color may indeed be the most discriminative feature. We show sample images of CUB in Fig.~\ref{fig:overview_dataset} (a).

As for iLab (Fig.~\ref{fig:overview_dataset} (b)), its task-specific bias is so even that we cannot reach a clear conclusion about which feature is more important. So we implement the class-specific biased experiments on iLab and summarize the class biases in Table \ref{table:ilab-local-bias}. It is interesting to find that the dominant feature is different for different classes. Take the boat as an example, it is strongly shape-biased, as indeed the shape of boats is so distinct in this dataset of vehicles. The results for military vehicles (mil) also deserve attention. The aggregation module shows that we can rely on color to distinguish mil. This conclusion is intuitive as the color of the mil is always green. We show more example images of iLab in the appendix.



\section{More Humanoid Applications with HVE}
To further explore more applications with HVE, we use HVE to simulate the visual reasoning process of humans and propose a new solution for conducting open-world zero-shot learning without predefined attribute labels (Sec.~\ref{sec:5.1}). We also use HVE to simulate human imagination ability through cross-feature retrieval and imagination (Sec.~\ref{sec:5.2}).

\subsection{Open-world Zero-shot Learning with HVE}
\label{sec:5.1}
\begin{figure*}[t]
\begin{center}
\includegraphics[width=\linewidth]{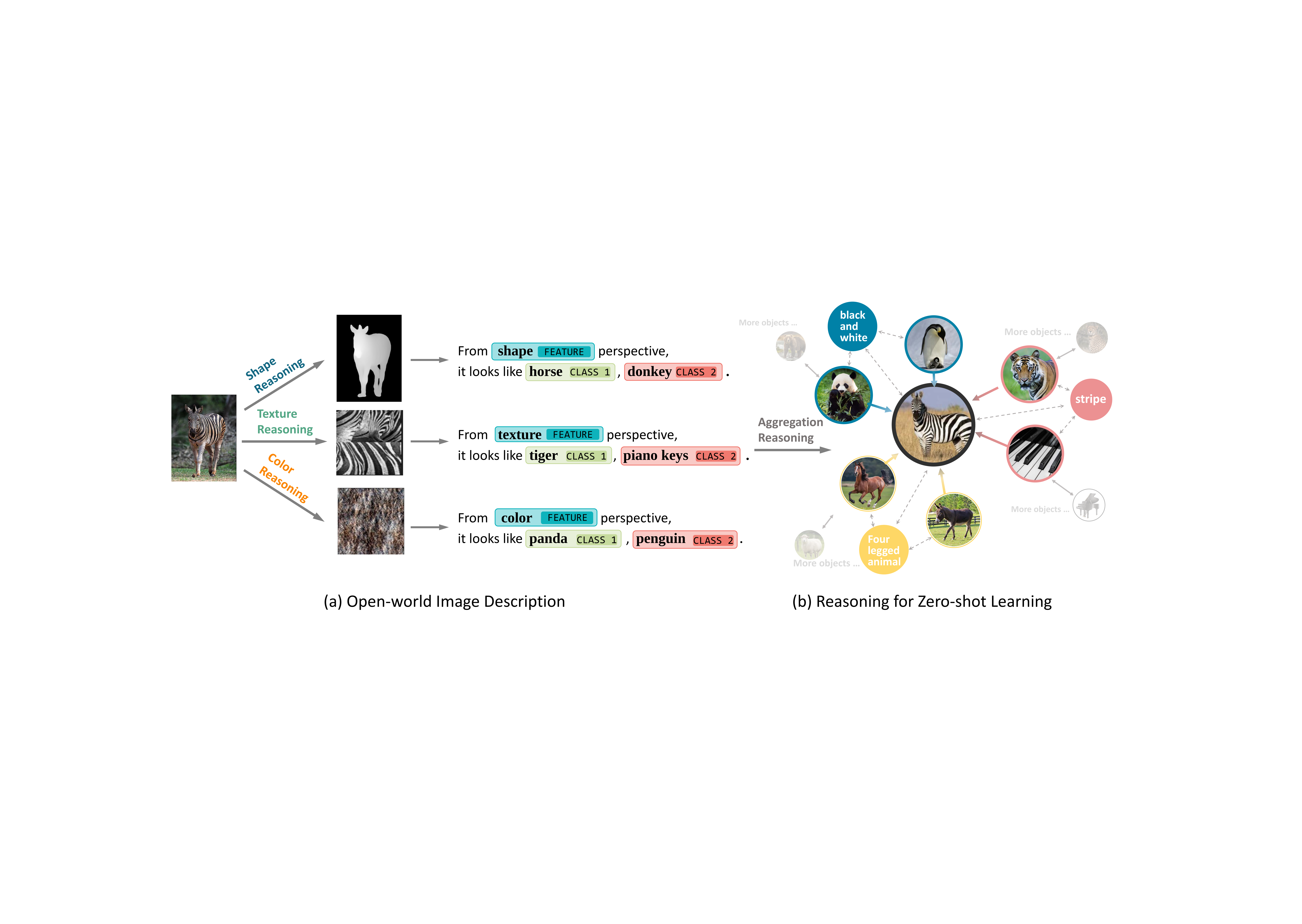}
\end{center}
  \caption{The zero-shot learning method with HVE. We first describe the novel image in the perspective of shape, texture, and color. Then we use ConceptNet as common knowledge to reason and predict the label.}
\label{fig:zero-shot}
\end{figure*}

Zero-shot learning is a challenging task, where, at test time, a learner needs to classify samples from classes never seen during training. Most current methods~\cite{PalatucciPHM09,lampert2009learning,fu2018recent} need humans to provide detailed attribute labels for each image, which is costly in time and energy. 
However, given an image from an unseen class, humans can still \textit{describe} it with their learned knowledge. For example, we may use horse-like shape, panda-like color, and tiger-like texture to describe an unseen class zebra, if we know horse, panda, and tiger.
We use HVE to simulate this feature-wise open-world image description by feature retrieval and ranking (Sec.~\ref{sec:5.1.1}).   
Based on the image description, humans may conduct further \textit{reasoning or consulting} with the help of an outside knowledge pool (e.g., WordNet~\cite{miller1995wordnet}) to predict the class of the given unseen image. 
Similarly, we propose a feature-wise open-world zero-shot learning (reasoning) pipeline with the help of ConceptNet~\cite{speer2017conceptnet} (Sec.~\ref{sec:5.1.2}).
The whole process can be seen in Fig.~\ref{fig:zero-shot}.


\subsubsection{Step 1: Description}
\label{sec:5.1.1}

In this section, we introduce how to use HVE to provide feature-wise description for any unseen class image without predefined attribute labels. First, to represent learnt knowledge, we use feature extractors $\mathnormal{E}_s,\mathnormal{E}_t,\mathnormal{E}_c$ (described in Sec.~\ref{section:HNN}) to get the shape, texture, and color representation $V_s^{(k,i)},V_t^{(k,i)},V_c^{(k,i)}$ of the $i_{th}$ image of seen class $k$. Then, given an unseen class image $I_{un}$, we use the same feature extractors to get its feature-wise representation $V_s^{\prime},V_t^{\prime},V_c^{\prime}$. To retrieve learnt classes as description, we calculate the average distance $d_m^{k}$ between $I_{un}$ and images of other class $k$ in the latent space on feature $m$. Here $m\in\{s,t,c\}$ is ``shape", ``texture" or ``color".
\begin{align}
&d_m^{k} = \frac{1}{n_k}\sum_{i\in \mathcal{T}_k}d_m^{(k,i)} = \frac{1}{n_k} \sum_{i\in \mathcal{T}_k}||V_m^{\prime}-V_m^{(k, i)}||_2,
\end{align}
where $\mathcal{T}_k$ represents images of seen class $k$ in the training set and $n_k$ is the number of images in class $k$. 

In this way, we can find the top $K$ closest classes of $I_{un}$ from shape perspective, and we call these $K$ classes shape roots $R_s$. We do the same operation to texture and color representation and get $R_t,R_c$. Now, we can describe $I_{un}$ using our roots. For example, as shown in Fig.~\ref{fig:zero-shot}(a), for the unseen class zebra, $R_s=\{\text{horse}, \text{donkey}\}$, $R_t=\{\text{tiger}, \text{piano keys}\}$, $R_c=\{\text{panda}, \text{penguin}\}$, and we can describe its shape, texture, and color based on these classes.

\subsubsection{Step 2: Open-world classification}
\label{sec:5.1.2}

To further predict the actual class of $I_{un}$ based on the feature-wise description,
we use ConceptNet as common knowledge to conduct reasoning.
As shown in Fig.~\ref{fig:zero-shot}(b), 
for every feature root $R_s,R_t,R_c$, we retrieve their common attribute as $R_a^m = \{\bigcap_{r\in R_m}^{}p_r\}$ in ConceptNet, where $p_r$ is the neighbors of $r$ in ConceptNet, (e.g., stripe is common attribute root of $R_t=\{\text{tiger}, \text{piano keys}\}$).
We form a reasoning root pool $R^{*}$ consisting of feature roots $R_s,R_t,R_c$ obtained during image description, and shared attribute roots $R_a^s$, $R_a^t$, $R_a^c$.
The reasoning roots will be our \textit{evidence} for reasoning.
For every root $r\in R^{*}$, we can search its neighbors in ConceptNet, which are treated as possible candidate classes for $I_{un}$. 
All candidates form a possible candidate pool $P$, which contains all hypothesis classes. 
Now we have two pools, root pool $R^{*}$ and candidate pool $P$.
For every candidate $p_i \in P$, we calculate the ranking score of $p_i$ as:
\begin{align}
    \Bar{S}(p_i) = \sum_{r_j\in R^{*}}^{}\cos(\mathcal{E}(p_i),
    \mathcal{E}(r_j)),
\end{align}
where $\mathcal{E}(\cdot)$ is the word embedding in ConceptNet and $\cos(A,B)$ means cosine similarity between $A$ and $B$. 

We choose the candidate having the highest-ranking score as our predicted label. We build a prototype zero-shot learning dataset with 34 seen classes as the training set and 5 unseen classes as the test set, using 200 images per class. We calculate the accuracy of unseen classes (Table~\ref{table:zero-shot}). Although not perfect, results are well above the chance level (1/5=20\%). The performance of each feature extractor and encoder in HVE is the core of open-world classification applications.  Fore baseline method, we conduct prototypical networks \cite{snell2017prototypical} on the same open-world zero-shot learning task with the same dataset and same backbone as our method. It is hard for the prototypical network to direct conduct zero-shot so we provide one-shot for each test class and use its one-shot setting. More details are in the appendix.

\begin{table}
\caption{Accuracy of unseen class for zero-shot learning}
\begin{center}
\begin{tabular}{c|c|c|c|c|c}
\hline
Method & fowl & zebra & wolf & sheep & apple  \\
\hline
Prototype (one-shot) & 19\% & 16\% & 17\% & 21\% & 74\%\\
Ours (zero-shot) & 78\% & 87\% & 63\% & 72\% & 98\%\\
\hline

\end{tabular}
\end{center}

\label{table:zero-shot}
\end{table}

\subsection{Cross Feature Imagination with HVE}
\label{sec:5.2}
We show HVE has the potential to simulate human imagination ability. We humans can intuitively imagine an object when seeing one aspect of a feature, especially when this feature is prototypical (contribute most to classification). For instance, we can imagine a zebra when seeing its stripe (texture). This process is similar but harder than the classical image generation task since the input features modality here  {\em dynamic} which can be any feature among shape, texture, or color. To solve this problem, using HVE, we separate this procedure into two steps: (1) \textbf{cross feature retrieval} and (2) \textbf{cross feature imagination}. Given any feature (shape, texture, or color) as input, cross-feature retrieval focuses on finding the most possible two other features. Cross-feature imagination can then generate a whole object based on a group of shapes, textures, and color features.

\subsubsection{Cross Feature Retrieval}

In order to reasonably retrieve the most possible other two corresponding features given only one feature (among shape, texture, or color), we learn a feature agnostic encoder that projects the three features into one same feature space and makes sure that the features belonging to the same class are in the nearby regions. 

\begin{figure}[h]
\begin{center}
\includegraphics[width=\linewidth]{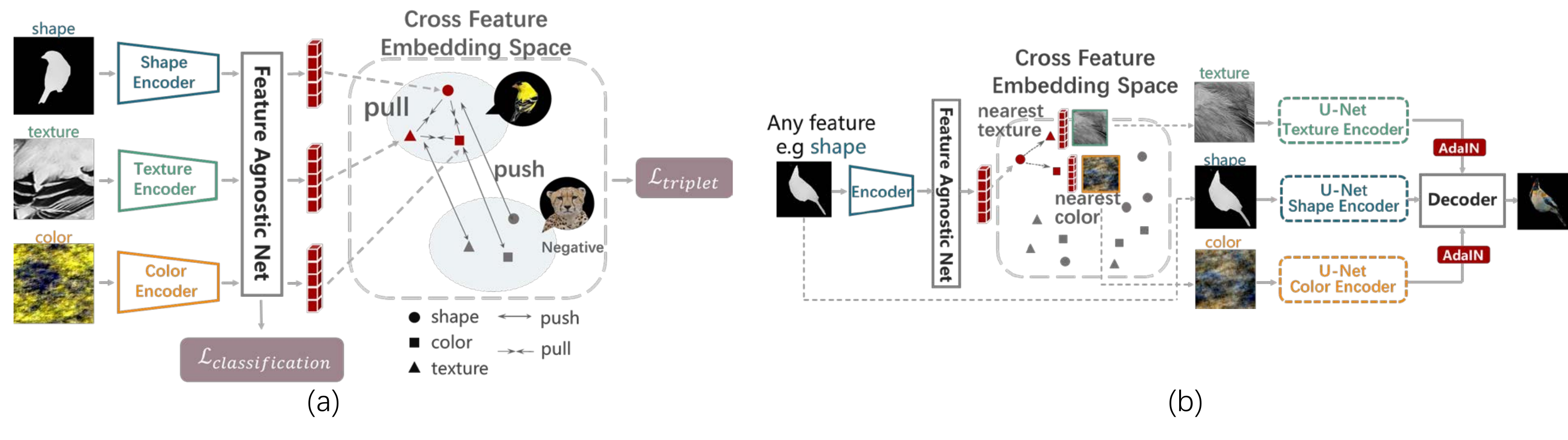}
\end{center}
   \caption{(a) The structure and training process of the cross-feature retrieval model. $E_{s}$, $E_{t}$, $E_{c}$ are the same encoders in Sec.~\ref{section:HNN}. The feature agnostic net then projects them to shared feature space for retrieval. (b) The process of cross-feature imagination. After retrieval, we design a cross-feature pixel2pixel GAN model to generate the final image.}
\label{fig5.1}
\end{figure}

\begin{table}
\setlength\tabcolsep{2pt}
  \scriptsize
  \caption{Cross-features retrieval accuracy on biased datasets (DS).}
  \centering
\begin{tabular}{c | c c  c | c c  c | c c  c }
    \toprule
      input & \multicolumn{3}{c|}{shape} & \multicolumn{3}{c|}{texture} & \multicolumn{3}{c}{color} \\ \hline 
    retrieval & shape& texture & color & shape& texture & color & shape& texture & color\\
    \midrule \midrule
 shape biased DS  & 86\% & 81\% & 74\% & 76\% & 77\% & 66\%  & 64\% & 61\% & 60\%\\
  texture biased DS & 52\% & 51\% & 41\% & 67\% & 73\% & 63\%  & 52\% & 57\% & 54\%\\
   color biased DS & 59\% & 56\% & 54\% & 56\% & 61\% & 59\%  & 67\% & 75\% & 75\%\\
  \bottomrule
\end{tabular}

\label{table-2}
\end{table}

As shown in Fig.~\ref{fig5.1}(a), during training, the shape $I_s$, texture $I_t$ and color $I_c$ are first sent into the corresponding frozen encoders $E_{s}$, $E_{t}$, $E_{c}$, which are the same encoders in Sec.~\ref{section:HNN}. Then all of the outputs are projected into a cross-feature embedding space by a feature agnostic net $\mathcal{M}$, which contains three convolution layers. We also add a fully connected layer to predict the class labels of the features. 
We use cross-entropy loss $\mathcal{L}_{\text{cls}}$ to regularize the prediction label and a triplet loss $\mathcal{L}_{\text{triplet}}$~\cite{schroff2015facenet} to regularize the projection of $\mathcal{M}$. For any input feature $x$ (e.g., a bird A shape), 
positive sample $x_{\text{pos}}$ are either same class same modality (another bird A shape) or same class different feature modality (a bird A texture or color); negative sample $x_{\text{neg}}$ are any features from different class. 
$\mathcal{L}_{\text{triplet}}$ pulls the embedding of $x$ closer to that of the positive sample $x_{\text{pos}}$, and pushes it apart from the embedding of the negative sample $x_{\text{neg}}$. The triplet loss is defined as follow:
{\small
\begin{align}
    \mathcal{L}_{\text{triplet}} = \max(\Vert\mathcal{F}(x) - \mathcal{F}(x_{\text{pos}})\Vert_2 - \Vert\mathcal{F}(x) - \mathcal{F}(x_{\text{neg}})\Vert_2+\alpha,0),
\end{align}
}
where $\mathcal{F}(\cdot):=\mathcal{M}(E(\cdot))$, $E$ is one of the feature encoders. $\alpha$ is the margin size in the feature space between classes, $\Vert \cdot \Vert_2$ represents $\ell_2$ norm.

We test the retrieval model in all three biased datasets (Fig.~\ref{fig:dataset_tsne}) separately. In the retrieval process, given any feature of any object, we can map it into the cross feature embedding space by the  corresponding encoder net and the feature agnostic net. Then we apply the $\ell_2$ norm to find the other two features closest to the input one as output. The output is correct if they belong to the same class as the input.

For each dataset, we retrieve the three types of feature pair by pair, and the accuracy is shown in Table~\ref{table-2}. The retrieval performs better when the input feature is the dominant of the dataset, which again verifies the feature bias in each dataset.


\subsubsection{Cross Feature Imagination}

\begin{figure}[t]
\begin{center}
\includegraphics[width=\linewidth]{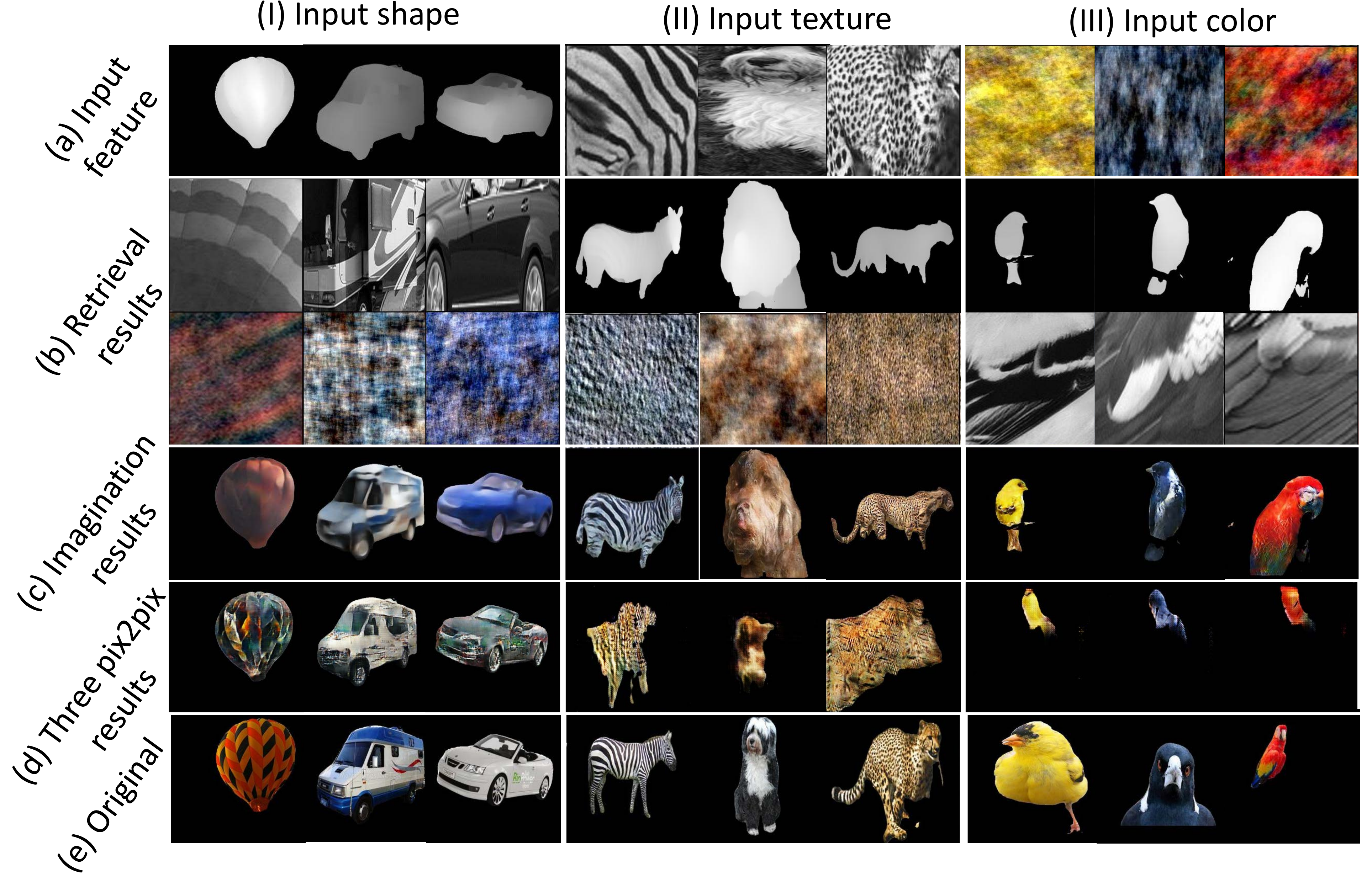}
\end{center}
\caption{Results of imagination with shape, texture, and color feature input (columns I, II, III). Line (a) is the input feature. Line (b) is the retrieved features given (a). Line (c) is the imagination results with HVE and our GAN model. Line (d) is results of baseline pix2pix GANs, which is only fed with input feature. Line (e) is the original images to which the input features belong. Our model can reasonably ``imagine" the object given a single feature.} 
\label{fig5.3}
\end{figure}

To stimulate imagination, we propose a cross-feature imagination model to generate a plausible final image with the input and retrieved features.
The procedure of imagination is shown in Fig.~\ref{fig5.1}(b). Inspired by the pixel2pixel GAN\cite{isola2017image} and AdaIN\cite{huang2017arbitrary} in the style transfer, we design a cross-feature pixel2pixel GAN model to generate the final image. The GAN model is trained and tested on the three biased datasets. In Fig.~\ref{fig5.3}, we show more results of the generation model. The results show that our model satisfyingly generates the object from a single feature. From the comparison between (c) and (d), we can clearly find that they are alike from the perspective of the corresponding input feature, but the imagination results preserve the feature from the retrieval features. The imagination variance also shows the
feature contributions from a generative view: if the given feature is the dominant feature of a class (contribute most in classification. e.g., the stripe of zebra), then the retrieved features and imagined images have smaller variance (most are zebras); While non-dominant given feature (shape of zebra) lead to large imagination variance (can be any horse-like animals). 
We create a baseline by using three pix2pix GANs (same structure as ours), where each pix2pix GAN is responsible for one specific feature (e.g., take one modality of feature as input and imagine the raw image). FID comparison shows in Table.~\ref{table:fid}. More details of the GAN model are in the appendix.
\begin{table}
  \small
  \centering
\caption{Cross-features imagination quality comparison.}
\begin{tabular}{c|c|c|c}
\hline
FID ($\downarrow$) & shape input  & texture input &  color input \\
\hline
Three pix2pix GANs & 123.915  & 188.854 & 203.527  \\
Ours & \textbf{96.871}  & \textbf{105.921} & \textbf{52.846} \\
\hline
\end{tabular}
\label{table:fid}
\end{table}




\section{Conclusion}
To explore the task-specific contribution of shape, texture, and color features in human visual recognition, we propose a humanoid vision engine (HVE) that explicitly and separately computes these features from images and then aggregates them to support image classification. With the proposed contribution attribution method, given any task (dataset), HVE can summarize and rank-order the task-specific contributions of the three features to object recognition. We use human experiments to show that HVE has a similar feature contribution to humans on specific tasks. We show that HVE can help simulate more complex and humanoid abilities (e.g., open-world zero-shot learning and cross-feature imagination) with promising performance. HVE is not a perfect simulation of HVS due to the limited learning ability and complexity of each component. These results are the first step towards better understanding the contributions of object features to classification, zero-shot learning, imagination, and beyond.

\textbf{Acknowledgments} This work was supported by C-BRIC (one of six centers in JUMP, a
Semiconductor Research Corporation (SRC) program sponsored by DARPA),
DARPA (HR00112190134) and the Army Research Office (W911NF2020053). The
authors affirm that the views expressed herein are solely their own, and
do not represent the views of the United States government or any agency
thereof.



\clearpage
%
%
\bibliographystyle{splncs04}
\bibliography{mian}

\clearpage
\appendix
\section*{Appendix}
\section{Details of Human Vision Engine (HVE)}
\subsection{Image Parsing and Foreground Identification}
\begin{figure}
\begin{center}
\includegraphics[width=0.85\linewidth]{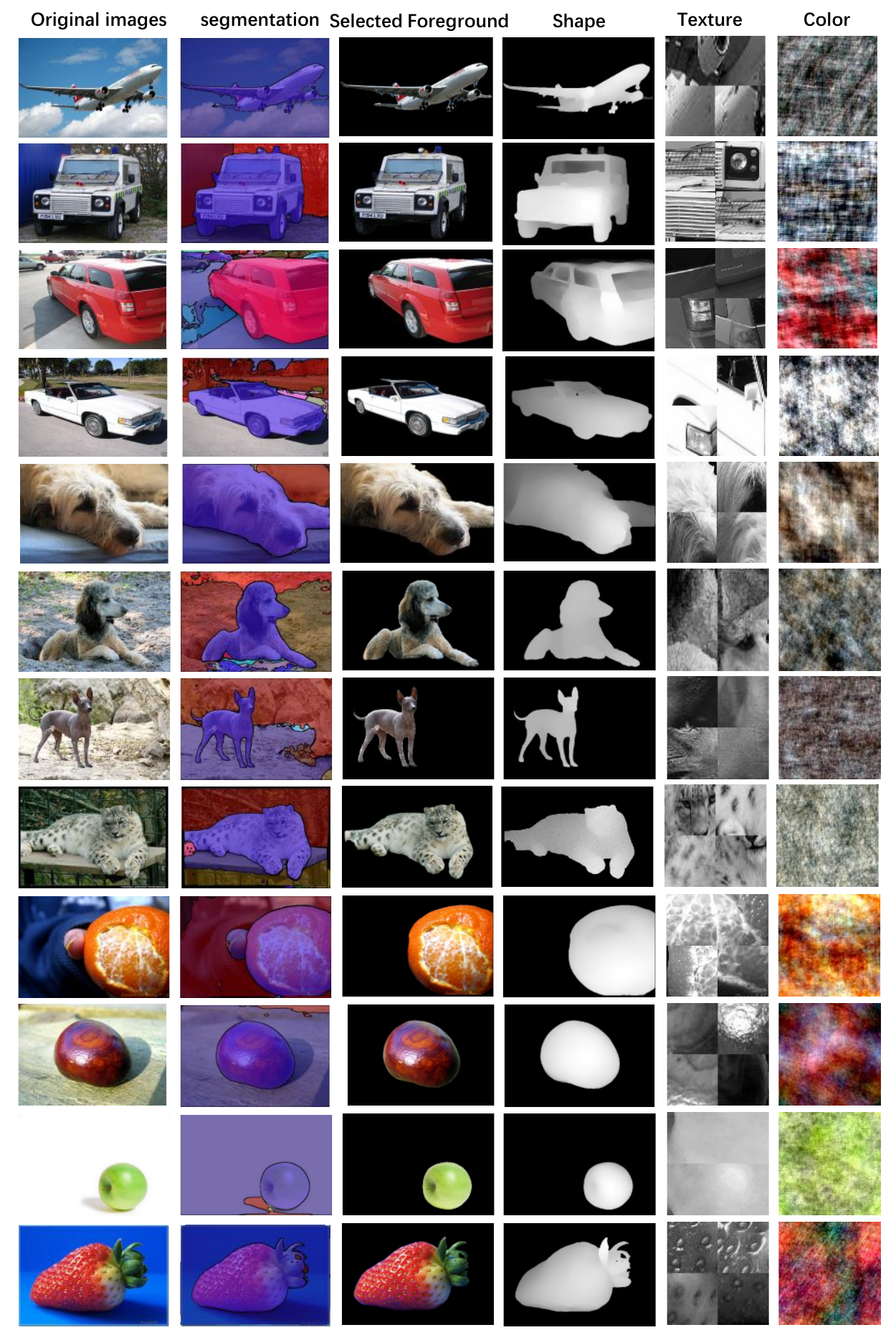}
\end{center}
\caption{Example of the preprocessing result and feature extraction.}
\label{fig:supp-feature-img}
\end{figure}

As described in the main paper Sec. 3.1, we use entity segmentation and foreground object identification to simulate the preprocessing behavior of the human vision system. An illustration is shown in Fig.~\ref{fig:supp-feature-img}.

The entity segmentation can parse an input image and output a set of binary images masks. Each mask represents a single object or stuff in the image. For an input raw image $I_{\text{raw}}\in \mathbb{R}^{H\times W\times C}$, we denote the image mask sets as $\{I_{\text{mask}, k}\}$, where $k=1,2,..., $ denotes each different object. For each image mask $I_{\text{mask},k}$, pixel $I_{\text{mask},k}^{(i,j)}$ equals to $1$ if and only if pixel $I_{\text{raw}}^{(i,j)}$ belongs to objects $k$, otherwise equals to $0$.

For foreground identification, we borrow the learned knowledge from a pretrained model which already learn the foreground class. As Grad-CAM \cite{selvaraju2017grad} could generate class-specific saliency map $M_{\text{cam}}$ given the model's prediction, which represent how important each pixel contribute to the specific prediction. So the pixel with a higher activation value is more likely belong to foreground. We first use a pretrained model to generate a saliency map $M_{\text{cam}}$ with Grad-Cam, and then we can get the binary attention map $M_{\text{att}}$ based on $M_{\text{cam}}$. For mask $M_{\text{att}}$, pixel $M_{\text{att}}^{(i,j)}$ equals to $1$ if and only if pixel $M_{\text{cam}}^{(i,j)} > \tau$ (we set $\tau$ to be the median value of $M_{\text{cam}}$), otherwise equals to $0$. 

After getting $M_{\text{att}}$, we can compute Intersection over union (IoU) score of each image mask $I_{\text{mask},k}$, 
$$S_k=SUM(I_{\text{mask},k}\cap M_{\text{att}})/SUM(I_{\text{mask},k})$$ 
Where $SUM()$ calculates the number of pixels. 
Then we choose the one with the highest score as the foreground mask.

\begin{figure}
\begin{center}
\includegraphics[width=0.9\linewidth]{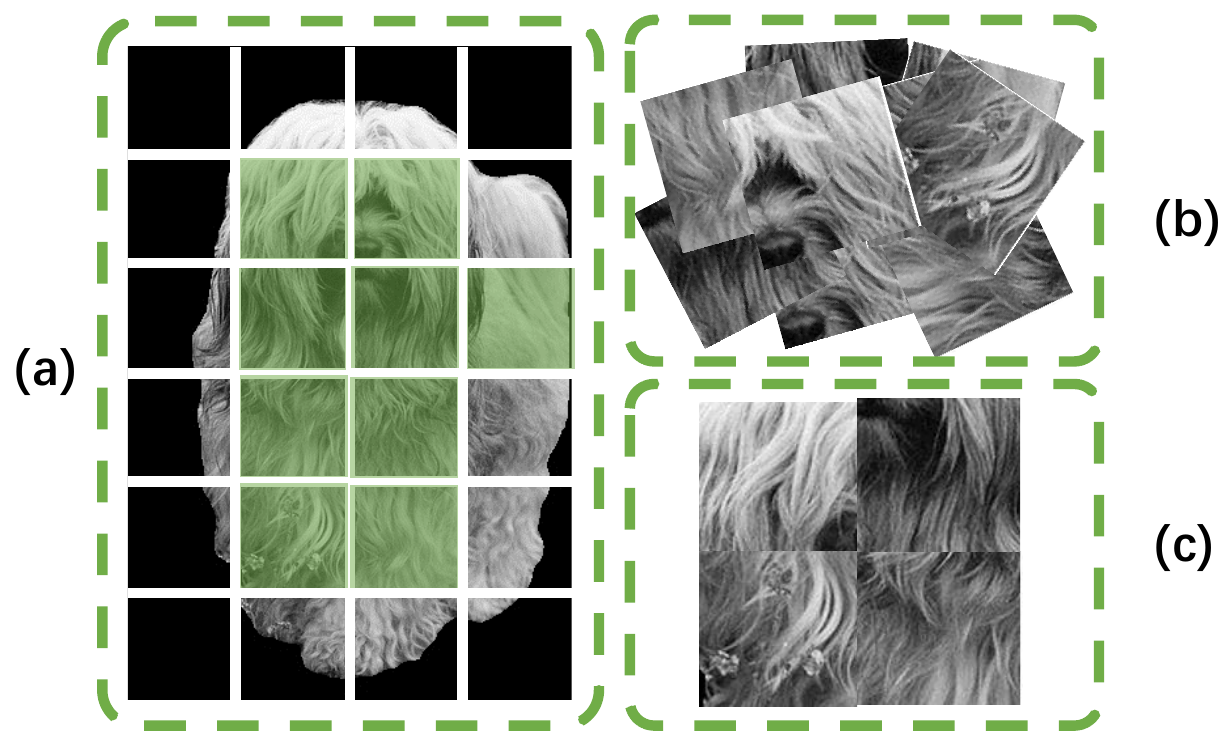}
\end{center}

\caption{The process for extracting texture. We cut the image of the foreground object into several square patches as shown in (a) and select the patch pool as shown in (b). (c) is the final texture feature, the concatenation of $k$ randomly selected patches from the patches pool.}
\label{fig:supp-texure}

\end{figure}


\subsection{Feature Extractor}

In HVE, with the preprocessed image $I\in\mathbb{R}^{H\times W\times C}$, we use the three independent feature extractors to obtain the corresponding image feature: shape image ($I_{s}$), texture image ($I_{t}$), color image ($I_{c}$). Here, we will introduce more details about the texture and color extractor.

\subsubsection{Texture Extractor}
\label{Sec.texture}
Fig.~\ref{fig:supp-texure} visualizes the process of extracting texture. First, to remove the color information, we convert the RGB object segmentation to a grayscale image. Since we want to get rid of the influence of background and extract the local texture feature as well as the global texture feature, we compute the  maximum circumscribed rectangle of the object by its 2D mask and resize this rectangle part to get a $224\times224$ image. In this way, we can eliminate the most background and focus on the object. Next, we will cut this new image into several square patches. Specifically, We first cut this image into several square patches, as shown in Fig.~\ref{fig:supp-texure} (a), if the overlap ratio between the patch and the original 2D object segment is larger than a threshold $\tau$, we will add them to a patch pool (we set $\tau$ to be 0.99 in our experiments, which means the over 99\% area of the patch belong to the object), as shown in Fig.~\ref{fig:supp-texure} (b). Since we want to extract both local texture information (one patch) and global texture information (whole image), we randomly select $k$ patches from the patch pool and concatenate them to a new texture image ($I_t$). We set $k=4$ because we want to concatenate those patches to a square image, which means $k$ should be a square number. If we set $k=1$, we can get the maximum square patch of the object but we lose some small, local texture. If we set $k=9$, the patch will be too small to contain useful information. To dynamically fit the object size and minimize the texture information loss. The number of patches we cut from the original image is dynamic. We will first cut the image into 9 square patches, if we can get $k=4$ valid patches that can be added into the patch pool from these 9 patches, we will stop processing this image. However, if we cannot get 4 valid patches, we will cut the original images into 16, 25, 36... smaller patches until we can get 4 valid patches. After concatenation, we can get the texture image $I_t$, such as Fig.~\ref{fig:supp-texure} (c). More results are shown in Fig.~\ref{fig:supp-feature-img}.



\begin{figure}
\begin{center}
\includegraphics[width=0.9\linewidth]{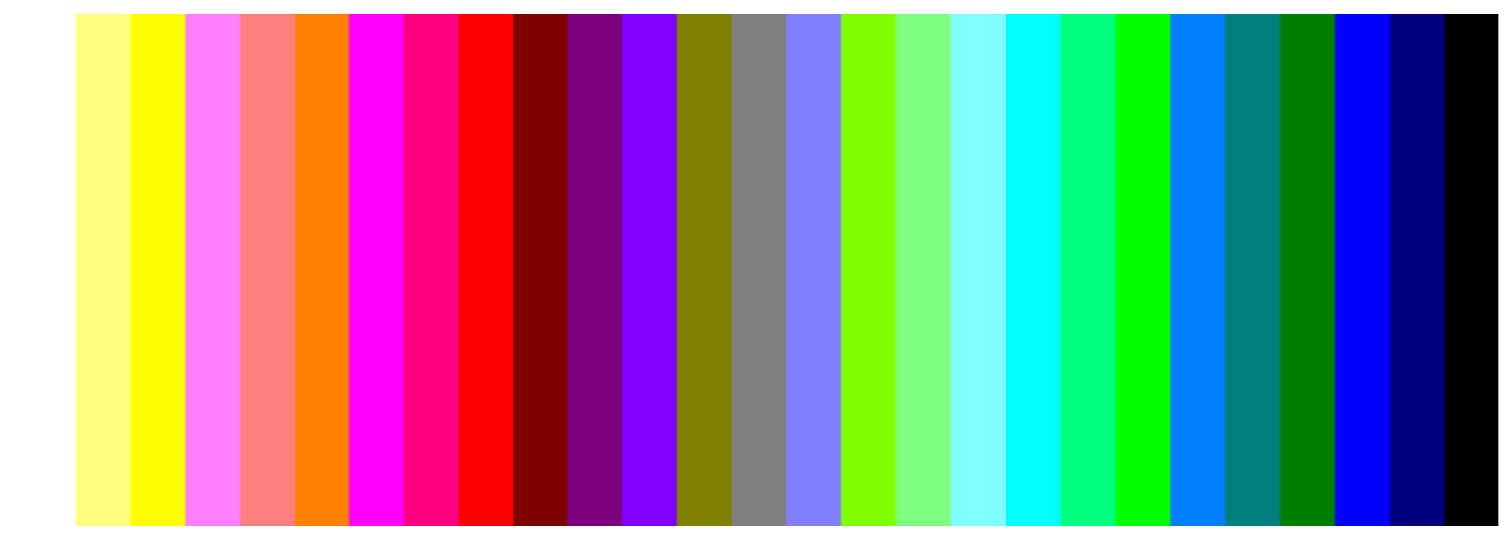}
\end{center}

\caption{27 Color centers for the color block.}
\label{fig:supp_center_img}
\end{figure}

\begin{figure}
\begin{center}
\includegraphics[width=\linewidth]{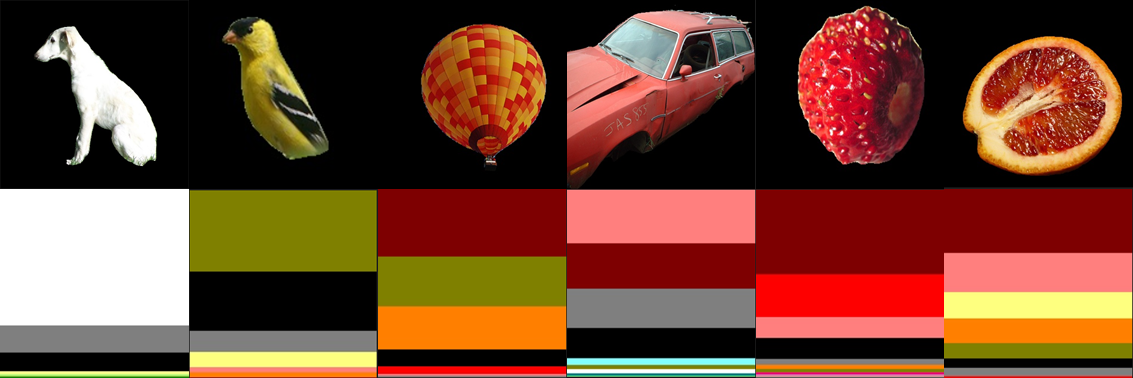}
\end{center}
\caption{Example of color block representation. The first row is original images, while the second row is the color blocks of those images.}
\label{fig:supp-color-block}
\end{figure}

\subsubsection{Color Extractor}
We have two different ways to extract the color feature $I_c$, phase scrambling, and color blocks. 
\vspace{-10pt}
\paragraph{Phase scrambling}
For a given image $I\in\mathbb{R}^{H\times W \times C}$ and a random matrix $z\in\mathbb{R}^{H\times W}$, we use the 2D fast fourier transform (FFT) on channel $j$ of the image and get the output $I_j^{\ast}\in\mathbb{C}^{H\times W}$, where $j=1,2,...,C$. We then caculate their modulus $r_j$ and angle $\theta_j$. Similarly, we apply the FTT to the random matrix $z$ and get the transformed result $z^{\ast}\in\mathbb{C}^{H\times W}$ and its angle $\varphi$. With $r_j, \theta_j, \varphi$, we can construct a new complex variable $T_j = r\cdot e^{i(\theta +p\varphi)}$, where $p\in[0,1]$ is a scramble factor. After the mapping backing by 2D inverse fast fourier transform and rescale the result to range $[0,255]$, the output will be the channel $j$ of our color feature $I_c\in\mathbb{R}^{H\times W \times C}$. This process can be described formally as:
 
\begin{align}
    I_j^{\ast} &= \text{FFT}(I_j) = r_i\cdot e^{i\theta_j}\\
    z^{\ast} &= \text{FFT}(z) = s\cdot e^{i\varphi},\\
    T_j &= r\cdot e^{i(\theta_j +p\varphi)},\\
    I_{c,j} &= \text{rescale}(\text{IFFT}(T_j)),
\end{align}
where $j=1,2,...,C$ and $I_{c,j}$ is the channel $j$ of $I_c$. More results are shown in Fig.~\ref{fig:supp-feature-img}.

\paragraph{Color Blocks}
The second method uses statistical color histogram representation \cite{van2009learning,gao2015hybrid}. The original RGB color space is a three dimensional cubic $\{(r, g, b) \lvert r, g, b \in [0,255]\}$. In order to represent the distribution of color for each image, we first choose $27$ center points which are uniformly distributed in the entire color space. The colors we choose are shown in Fig.~\ref{fig:supp_center_img}. For an input image, we assign each pixel to its closest center point by calculating their manhattan distance. By counting how many pixels belong to each color center and calculating the percentage, we can summarize the result to a color block image of size $224\times224$ as our color feature $I_c$ (the examples are shown in Fig.~\ref{fig:supp-color-block}). The color block consists of various stripes in different widths and colors. The color of stripes comes from the color center and the width represents how many percent of pixels this center covered in the input image. For instance, if there are 10\% pixels assign to white and 90\% pixels of black, we will generate a  image whose 10\% pixels are RGB $(255,255,255)$ and 90\% pixels are RGB $(0,0,0)$. This image ($I_c$) will not contain any shape or texture information. Fig.~\ref{fig:supp-color-block} shows some examples of the color blocks.

Compared with phase scrambling, this method is more intuitive to understand but may lose information when approximating RGB value in color space to color centers.




\subsection{Details about the Humanoid Neural Network}

\begin{table}[h]
\centering
\caption{Architecture of the humanoid neural network. $N$ is the number of labels, depending on which dataset is being used.}
\begin{tabular}{ccc}
\toprule\noalign{\smallskip}
Layer & Input $\rightarrow$ Output Shape & Layer Information\\
\noalign{\smallskip}
 \hline
 \hline
 \noalign{\smallskip}
 Shape Encoder & (224, 224, 1) $\rightarrow$ (7, 7, 512) & ResNet18  \\\noalign{\smallskip}
 Texture Encoder & (224, 224, 1) $\rightarrow$ (7, 7, 512) & ResNet18  \\\noalign{\smallskip}
 Color Encoder & (224, 224, 3) $\rightarrow$ (7, 7, 512) & ResNet18  \\\noalign{\smallskip}
 \midrule
 \noalign{\smallskip}
 Concatenation Layer & \ \  $3\times$(7, 7, 512) $\rightarrow$ (7, 7, 1536) & -  \\\noalign{\smallskip}
 Pooling Layer & \ \  (7, 7, 1536) $\rightarrow$ (1, 1, 1536)& Average Polling-($k$7x7, $s$1, $p$0)  \\\noalign{\smallskip} 
 Flatten Layer & \ \  ($1$,  $1$, 1536) $\rightarrow$ 1536 & - \\\noalign{\smallskip} 
\midrule
\noalign{\smallskip}
 Hidden Layer & \ \  1536 $\rightarrow$ 512 & \ \ \ Linear, ReLU \\\noalign{\smallskip}
\midrule\noalign{\smallskip}
 Output Layer & \ \  512 $\rightarrow$ $N$ & \ \ \ Linear, Softmax \\\noalign{\smallskip}
\bottomrule
\end{tabular}

\label{table:architecture}
\end{table}

The three encoders ($E_s, E_t, E_c$) use ResNet18 as backbone. $E_s$ takes shape feature with size $224 \times 224 \times 1$ as input, while $E_t$ takes texture feature with size $224 \times 224 \times 1$ and $E_c$ takes color feature with size $224 \times 224 \times 3$. They all produce the feature with size $7\times 7\times 512$. When training the encoders, these output features are then passed into a fully connected layer, which output the vectors with length $N$, the number of classes in the dataset.

During training the interpretable aggregation module, we freeze the three encoders and concatenate their output features
along the channel dimension into a tensor of size $7\times 7\times 1536$. This tensor is the input of the interpretable aggregation module, which is a two-layer MLP. The final output is a vector with length $N$, the number of classes in the dataset. The summary of the structure of the three encoders and the aggregation module is shown in Table~\ref{table:architecture}.


We use Adam optimizer with $\beta_{1} = 0.9$ and $\beta_{2}=0.999$, and set batch size to 64, learning rate is 0.001.

\section{Experiments Details}

\subsection{Influence of Random Selection in Texture Feature Extractor}

We re-ran 10 times main paper Table.2 using different random seeds for texture (patch selections and shuffling the order sequence). The mean and std shown in Table.~\ref{table:random}.

\begin{table}[h]
  \small
  \centering
\caption{Influence of Random Selection in Texture Feature Extractor.}
\begin{tabular}{c | c  c  c  c}
    \toprule
    contribution  & Shape & Texture & Color \\
    \midrule
     Shape-bias DS & 47.1\% $\pm$ 2.7\% & 34.5\% $\pm$ 2.2\% & 18.4\% $\pm$ 1.4\% \\
     Texture-bias DS & 5.2\% $\pm$ 1.8\% & 62.4\% $\pm$ 2.3\% & 32.3\% $\pm$ 1.6\% \\
     Color-bias DS & 12.4\% $\pm$ 2.1\% & 19.1\% $\pm$ 1.9\% & 68.5\% $\pm$ 3.6\% \\
     \bottomrule
\end{tabular}
\label{table:random}
\end{table}

\subsection{More quantitative contribution summary results of Humanoid NN}

To further evaluate the contribution of shape texture and color,
for each feature V (e.g., shape feature Vs), we compute the accuracy if we only use the rest features (e.g., combine texture Vt and color Vc) as input, and calculate the accuracy drop compared with using all three features as input (last column in Table 1). That accuracy drop represents the “unsubstitutability” or the “necessity” or “non-redundant contribution”  of that feature for visual recognition.  The results (Table.~\ref{table:two-feature}) are consistent with our previous results (main paper Table 1) on the contribution/importance of each feature. For example, in the shape biased dataset, the largest accuracy drop is when we remove the shape feature.

\begin{table}
\scriptsize
  \centering
\caption{More quantitative contribution summary results of Humanoid NN}
\begin{tabular}{c | c  c  c  c}
    \toprule
    accuracy  & Shape + Texture & Shape + Color & Texture + Color & \bf{all}\\
    \midrule
     Shape biased DS & 94\% & 92\% & 86\% & 95\%  \\
     Texture biased DS & 89\% & 80\% & 83\% & 90\% \\
     Color biased DS & 83\% & 88\% & 87\% & 92\%\\
     \bottomrule
\end{tabular}
\label{table:two-feature}
\end{table}

\subsection{Different interpretable aggregation module}
We conducted experiments to substitute the non-linear MLP with simple average pooling followed by an output classification layer. Contributions are shown in table.~\ref{table:aggre-exp}. While the numerical results differ, the ordering and conclusions remain (e.g., shape texture is most important in the shape-biased dataset). The experiments results show that the contribution result is robust to the aggregation module.
\begin{table}
  \small
  \centering
\caption{Average pooling interpretable aggregation module}
\begin{tabular}{c | c  c  c  c}
    \toprule
    Ave-Pool  & Shape ratio & Texture ratio & Color ratio \\
    \midrule
     Shape biased DS & 48\% & 40\% & 12\%   \\
     Texture biased DS & 1\% & 79\% & 20\%  \\
     Color biased DS & 1\% & 4\% & 95\% \\
     \bottomrule
\end{tabular}
\label{table:aggre-exp}
\end{table}

\subsection{Sample Question of Human Experiments}

\begin{figure}
\begin{center}
\includegraphics[width=\linewidth]{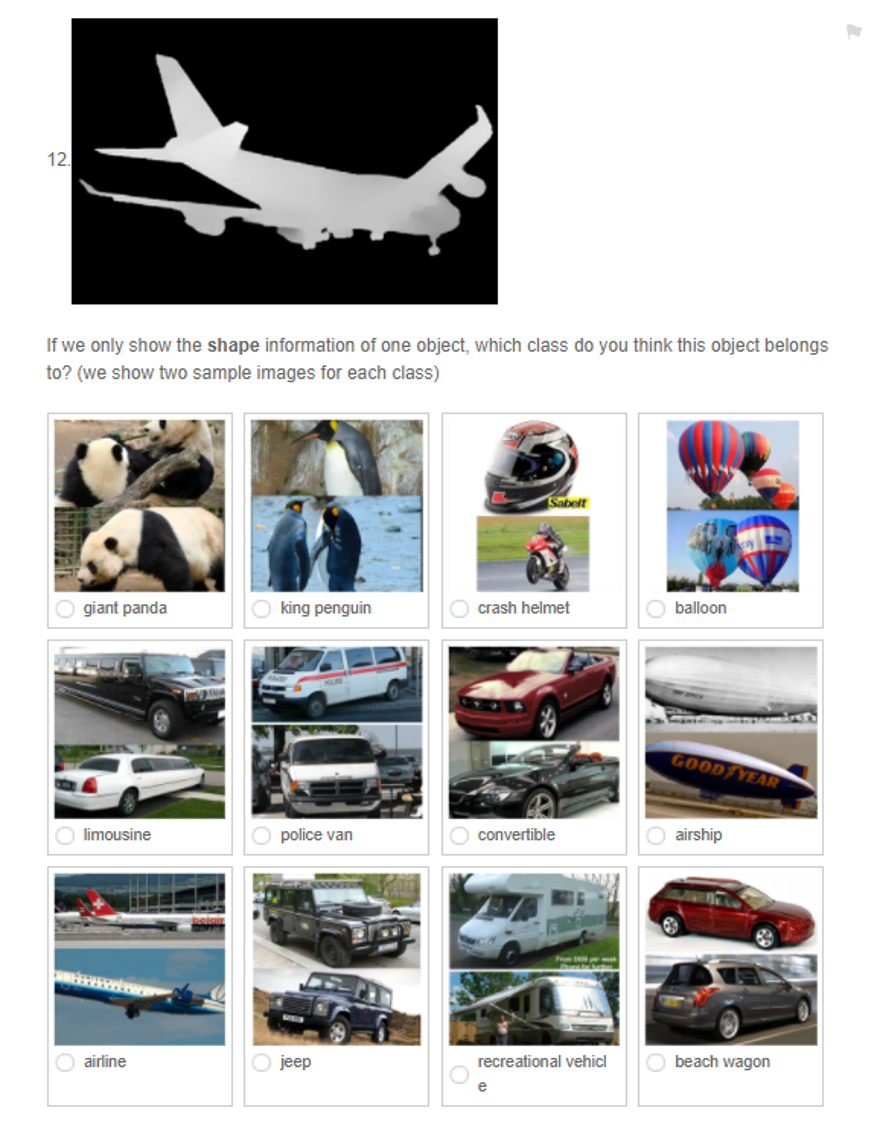}
\end{center}
   \caption{A screenshot of an experiment trial in our human experiments.}
\label{fig:supp-question}
\end{figure}

We designed human experiments that asked participants to classify reduced images with only shape, texture, or color features. Here, as shown in Fig.~\ref{fig:supp-question}, we demonstrate a screenshot of an experiment trial in our experiments.

\subsection{Contributions of Features in Different Tasks}

 
  

\begin{figure}
\begin{center}
\vspace{-10pt}
\includegraphics[width=0.55\linewidth]{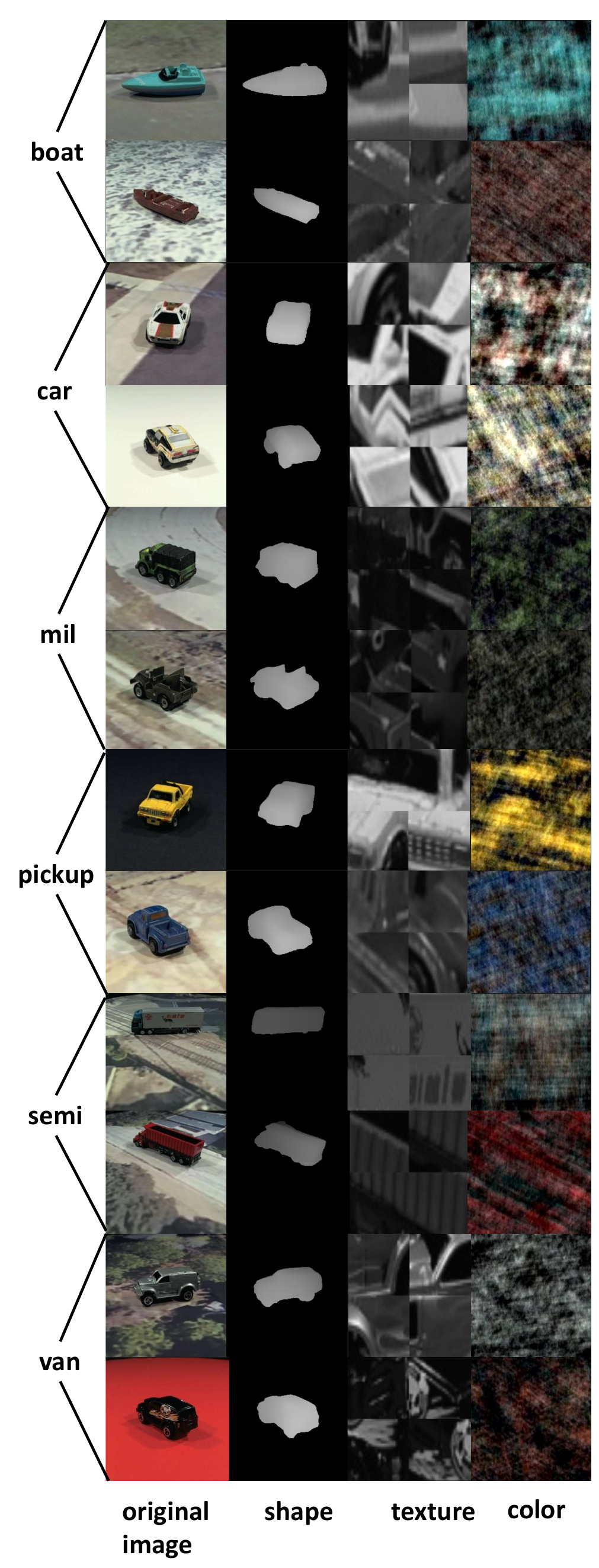}
\end{center}
\caption{Example images for ilab-20M classes.}
\label{fig:supp-ilab}
\end{figure}

\begin{figure}
\begin{center}
\includegraphics[width=\linewidth]{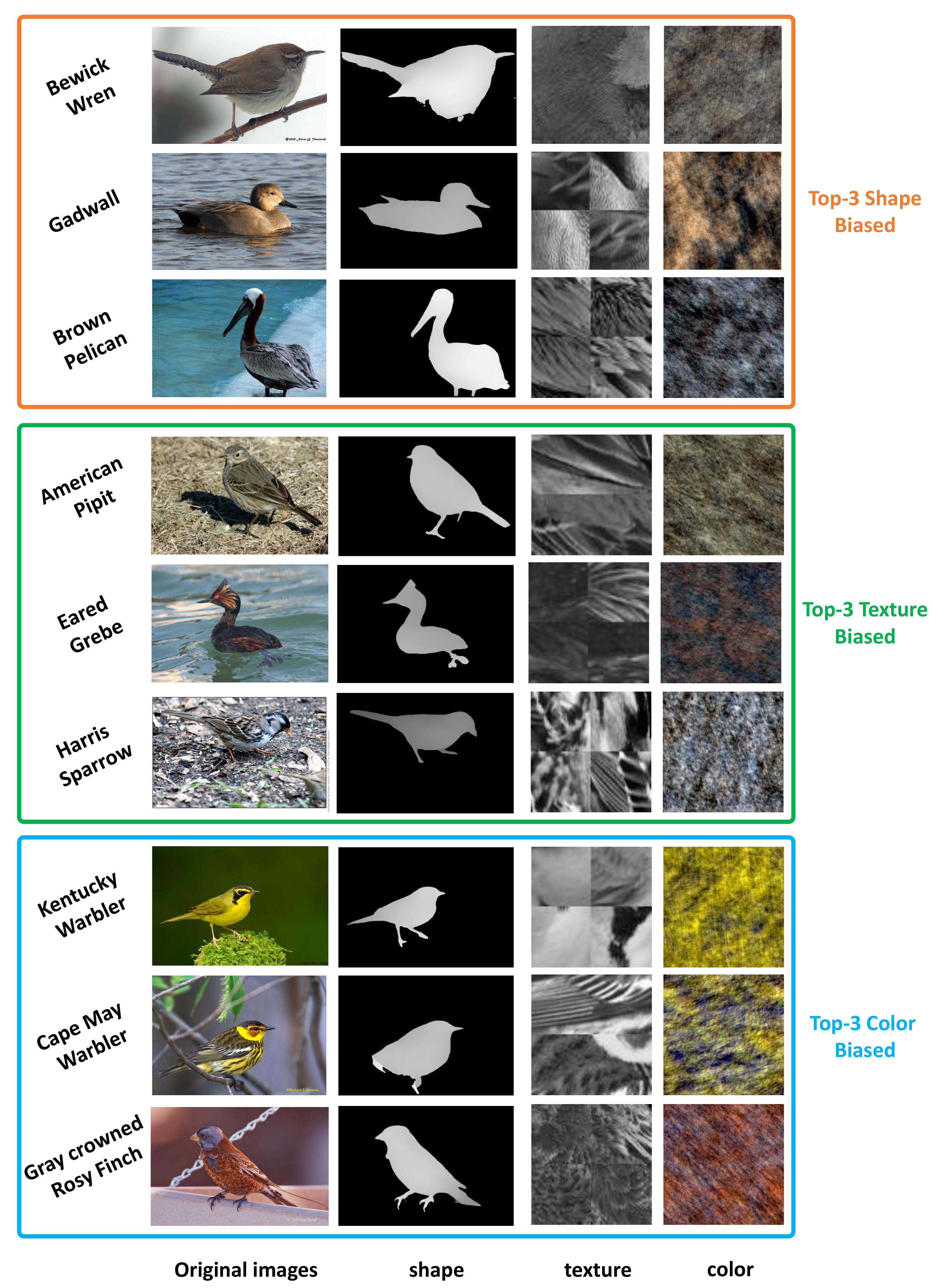}
\end{center}
\caption{Example images for CUB classes. The first 3 rows show the top-3 classes in CUB whose shape contribution is highest; The middle 3 rows show top-3 classes whose texture contribution is highest; The last 3 rows show top-3 classes whose color contribution is highest. Table.~\ref{table:CUB-local-bias} shows the quantity biased results for these classes.}
\label{fig:supp-CUB}
\end{figure}

\begin{table}
\setlength\tabcolsep{3pt}
  \footnotesize

  \centering
  \caption{Local bias for CUB classes.}
\begin{tabular}{l | c  c  c}
    \toprule
    ratio & shape & texture & color \\
    \midrule \midrule
Bewick Wren & \bf{44.8\%} & 21.3 & 33.8\%\\
Gadwall & \bf{43.6\%} & 24.2\% & 32.2\%\\
Brown Pelican & \bf{41.6\%} & 33.0\% & 25.4\%\\
American Pipit & 6.8\% & \bf{72.7\%} & 20.4\%\\
Eared Grebe & 5.6\% & \bf{70.3\%} & 24.1\% \\
Harris Sparrow & 13.5\% & \bf{64.1\%} & 22.4\% \\
Kentucky Warbler & 0.1\% & 0.7\% & \bf{99.2\%} \\
Cape May Warbler & 0.3\% & 1.2\% & \bf{98.5\%} \\
Gray crowned Rosy Finch &1.1\%  & 1.5\% & \bf{97.4\%}\\
  \bottomrule
\end{tabular}

\label{table:CUB-local-bias}
\end{table}

Fig.~\ref{fig:supp-ilab} shows more example images of the iLab dataset \cite{borji2016ilab}. In the figure, we can see that the military vehicles are always in the unique green, which matches with the result in our experiment that color is the most discriminative feature to classify military among other vehicles.

We also compute local bias for each class in CUB dataset \cite{wah2011caltech}. For each feature among shape, texture, and color, we select the top 3 classes whose feature contributions are the highest, and we show the bias for these 9 classes in Table.~\ref{table:CUB-local-bias}. To show more details of these top classes for each feature, we show the example images in Fig.~\ref{fig:supp-CUB}. For the first three-row classes, the shape is the most discriminative feature to classify them among other birds; for the middle three-row classes, the texture is the most discriminative feature, while color is the most discriminative feature to classify the last three-row classes.

\section{Details of Humanoid Application}

\subsection{Process of Open-world Zero-shot Learning}

\begin{table}[h]
\begin{center}
\begin{tabular}{c|c|c|c|c|c}
\hline
\textbf{Class} & zebra  & fowl & wolf & sheep & apple  \\
\hline
\hline
\textbf{shape root} & horse, donkey & \makecell[c]{turkey, goose\\cock} &\makecell[c]{ hyena, fox, \\tiger} & goat, bull & \makecell[c]{tomato, orange, \\pear}\\
\hline
\textbf{texture root} & \makecell[c]{tiger,\\ piano keys} & turkey, penguin & fox, dog & dog, goat & cherry, tomato\\
\hline
\textbf{color root} & panda, penguin & turkey & dog & \makecell[c]{elephant, \\goat }& cherry, tomato\\
\hline

\end{tabular}
\end{center}
\caption{More open-world zero-shot learning result.}
\label{table:more-zero-shot}
\end{table}

As describe in main paper Sec.5.1, we conduct open-world zero-shot learning with HVE. Here, we describe more details about the experiment. Given an unseen class image, we can use open world description (Step 1) to get its shape root, texture root, color root. For example, given a wolf image, we can get $R_s=\{\text{hyena}, \text{fox, \text{tiger}}\}$, $R_t=\{\text{fox}, \text{dog}\}$, $R_c=\{\text{dog}\}$. We provide more examples in table.\ref{table:more-zero-shot}. 

After we get $R_s=\{\text{hyena}, \text{fox, \text{tiger}}\}$, $R_t=\{\text{fox}, \text{dog}\}$, $R_c=\{\text{dog}\}$, we use ConceptNet \cite{speer2017conceptnet} and word embedding to predict its label (Step 2 in main paper Sec. 5.1).  Specifically, if we directly use the candidate pool got from ConceptNet and calculate their ranking score, the top 5 results are animal, four-legged animal, mammal, wolf, fox. We can find wolf are the first concrete class we can obtain, which achieved our open-world zero-shot classification. 

To show the quantitative results of our method, we realize that, although the first 3 results (animal, four-legged animal, mammal) in the previous wolf example is somewhat correct, we want to get a more concrete answer which is a wolf. 
Thus, we design a fixed candidate which excludes these broad words and only contains unseen classes and some disturbances. In this way, we can get a quantitative accuracy without the influence of those broad words like animal, fruit, bird.


Here, we provide the experiment setting of our seen class list and our candidate pool. Our seen classes list totally contain 36 classes, they are horse, tiger, panda, penguin, piano keys, cheetah, hyena, dog, lemon, koala, fur, squirrel, fox, rabbit, goose, sea lion, elephant, otter, duck, cock, chimpanzee, goat, orange, ball, bull, tomato, cherry, pear, turkey, seal, porpoise, alpaca, pigeon, lion, donkey, bear. Our candidate pool totally contains 20 classes which are fowl, zebra, bear, wolf, husky, swan, giraffe, jackal, peach, sheep, seal, apple, banana, train, bag, balloon, car, pen, table, eagle. The results are shown in the main paper Table 5.

\subsection{Detail of Prototypical network}
As described in the main paper Sec.5.1, we conduct one-shot learning using the prototypical network \cite{snell2017prototypical}. Here, we provide more details about the experiment. Prototypical network use the same training and test set with HVE. It is not clear for a prototypical networks to direct conduct zero-shot in the released code, so we provide a easier mode, one-shot for each test class, and use its one-shot setting. To train the prototypical network, we set the input image size as $3\times 224 \times 224$, hidden dimension as 64, learning rate as $1e-3$. 
We use their official code and follow the same training strategy as \cite{snell2017prototypical}. 
During testing, we use 5-way 1-shot setting,
table.5 in the main paper provide the final result of the prototypical network.



\subsection{The GAN model in Cross Feature Imagination}

\begin{figure}
\vspace{-10pt}
\begin{center}
\includegraphics[width=\linewidth]{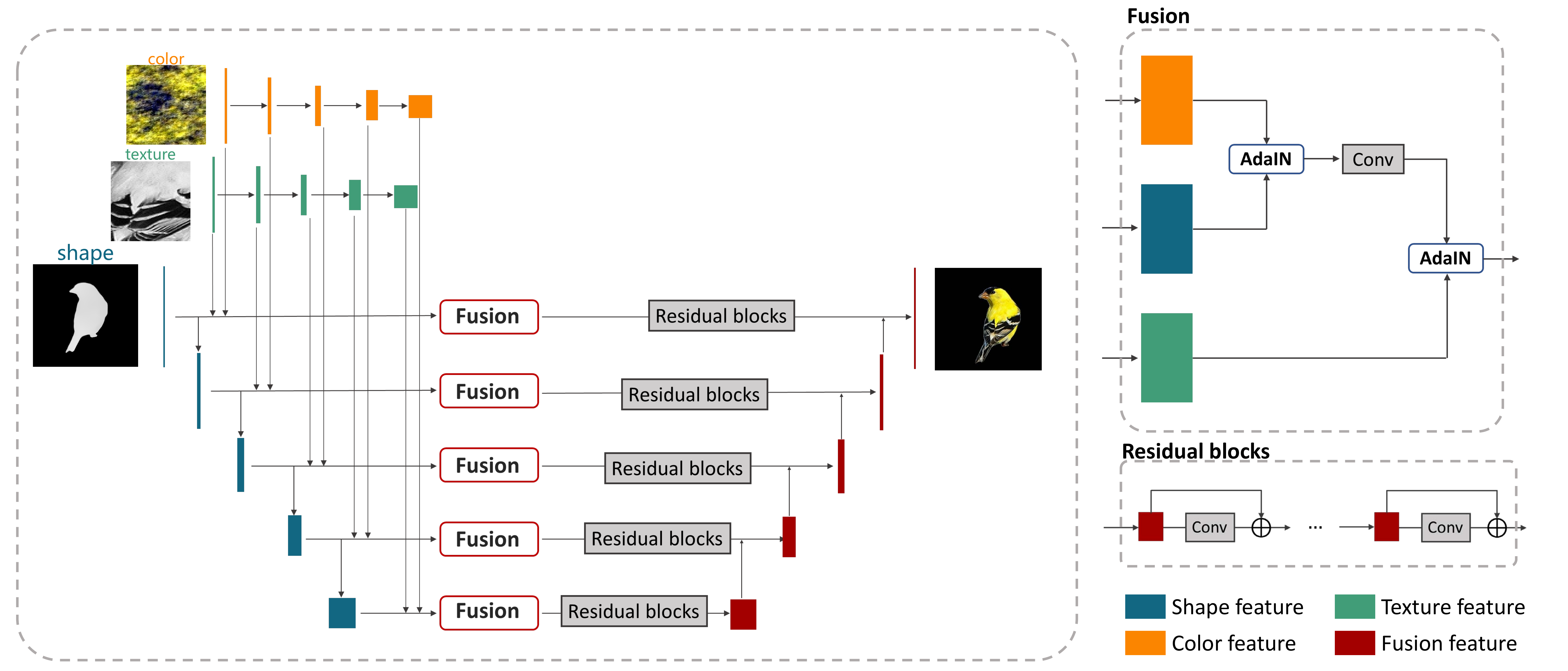}
\end{center}
  \caption{The architecture of the GAN generator in cross feature imagination.}
\label{fig:GAN-pipeline}
\end{figure}

As introduced in main paper Sec. 5.2, we design a cross-feature pixel2pixel GAN model to generate the final image. 

Fig.~\ref{fig:GAN-pipeline} shows the architecture of the generator. In order to ensure the quality of image generation, we set $k=1$ when extracting the texture feature (the selection of $k$ is introduced in Sec.~\ref{Sec.texture}). Each feature, shape, texture, and color are encoded by a sequence of convolutions maps to $F_{s,i},F_{t,i},F_{c,i}, i=1,2,...,K$. These features are fused into $F_i$ by $K$ feature fusion modules. As for fusion modules, we apply AdaIN and $L$ residual blocks to blend the different features. The outputs are de-convoluted into $H_i, i=1,2,...,K$, and then we get the final result. In our experiments, we use $K=5$ and $L=5$.

We use the same discriminator and loss function as \cite{isola2017image}. In training the GAN model. We use Adam optimizer with $\beta_{1} = 0.5$ and $\beta_{2}=0.999$ for both generator and discriminator, and set batch size to 16. The model is trained for 200 epochs on the training set, and we use the learning rate at $2e-4$ in the first 100 epochs and $2e$-5 from epoch 100 to 200. To compare our result, we use the original pix2pix GANs~\cite{} as our baseline model. The GAN model takes one type of feature as input and the original image as output. We trained three separate pix2pix GANs for each feature. 

\begin{figure}[h!]
\begin{center}
\includegraphics[width=\linewidth]{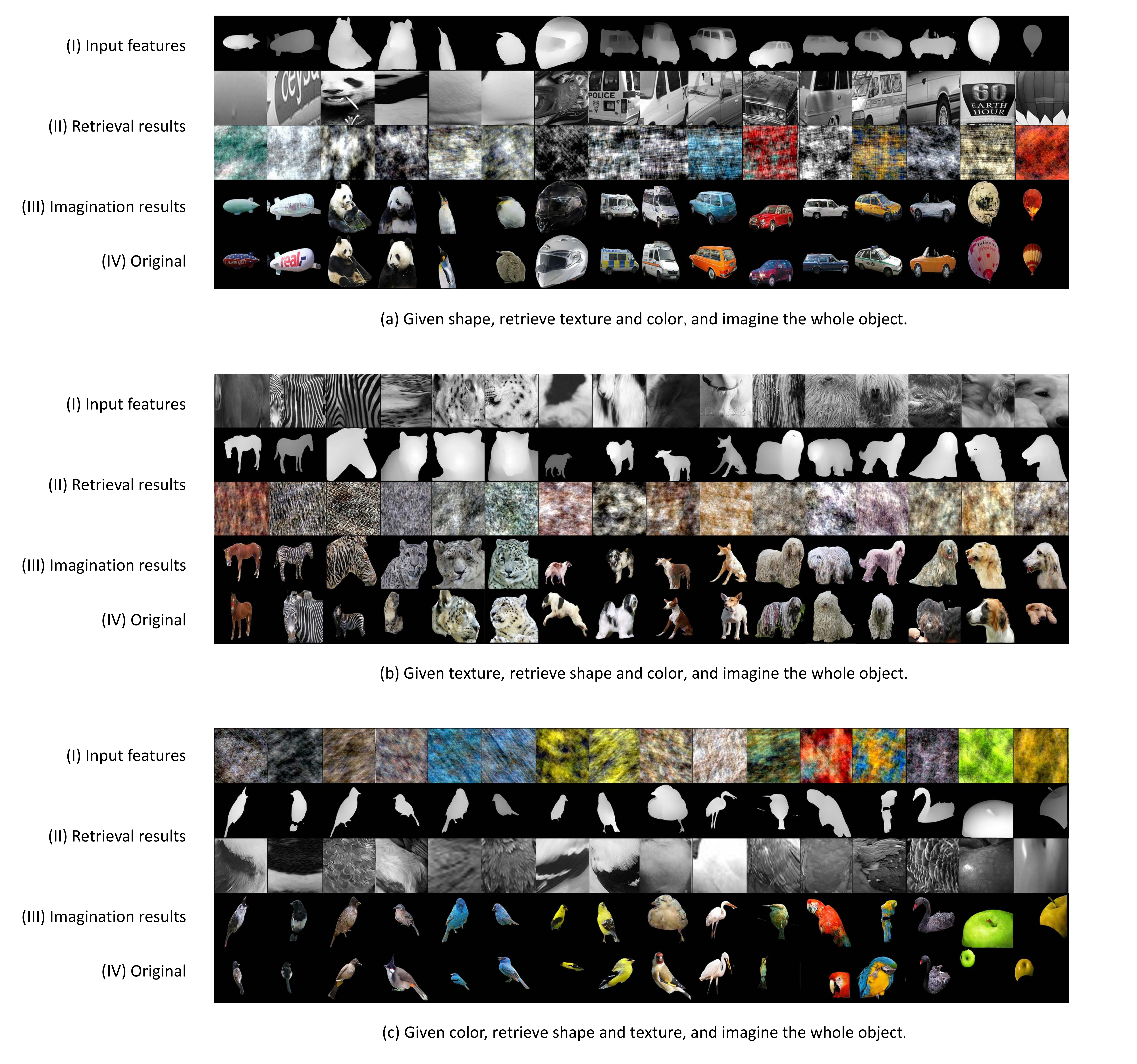}
\end{center}
   \caption{Results of the imagination when different types of features are given (see sub-images a, b, c). Given one feature (Line I), the retrieval model can find the other two plausible features (Line II), and then the GAN model can imagine the whole objects (Line III). Line IV shows the original images of the input features.}
\label{fig:GAN-result}
\end{figure}

In Fig.~\ref{fig:GAN-result}, we show more results about the imagination on the test set when different types of features are given (denoted as sub-image a, b, c). With one feature (Line I), the retrieval model can find the other two plausible features (Line II), and then the GAN model can imagine the whole objects (Line III). Comparing the imagination results and the original images to which the input features belong (Line-IV), we can find that they have a similar input feature, while the other two features of the imagination images are decided by the 
retrieved features.

\section{Discussion of Limitations and Future Work}

As we use Grad-Cam as the foundation for selecting foreground parts, sometimes models could output the wrong saliency map (for example, the pre-trained models may regard water as the most important object for classifying a boat. As a result, our image parsing pipeline would take water as the foreground). 
So how to select accurate foreground parts is the first step we need to improve in the future.

After getting the foreground part of images, we use three feature extractors to mimic the process of humans perceiving this world. Our feature extractors are far away from perfect. In this work, we concatenate four square patches to get $I_t$, but this introduced two lines in the $I_t$ (the line between different squares). These lines may cause confusion to the neural networks. What's more, 4 patches may not be the best choice to represent the object's texture. Some classes are too complex to be represented by only 4 patches. For example, cars have windows, metal, rubber, paint, and plastic. Different parts of cars could have different textures. 4 square patches can represent the texture of cars to some degree, but if we want to get the representation of all texture information without any omission, we need to improve the work process of the texture extractors. However, the main goal of our model design is to summarize the contributions by end-to-end learning, with minimal human-introduced bias and assumptions in the architecture design. We mainly provide the first fully objective, data-driven, and indeed first-order, measure of the respective contributions. 

In the open-world zero-shot learning experiment, we use a prototype dataset that contains limited classes, we will explore this direction with a larger dataset.

We think our HVE take a small but important step towards, from a humanoid perspective, better understanding
the contributions of shape, texture, color to classification, zero-shot learning, imagination, and beyond.

\paragraph{Dataset copyright.} We used publically available data. ImageNet, CUB, iLab-20M.

Imagenet license: Researcher shall use the Database only for non-commercial research and educational purposes
We use a subset of the ILSVRC2012 dataset (ImageNet) and iLab-20M. The object in iLab-20M is toy vehicles, under
Creative Commons CC-BY license. They do not contain any personally identifiable information offensive content.


\end{document}